\documentclass{article}
\usepackage{spconf,amsmath,amssymb,graphicx,booktabs,hyperref,microtype}
\usepackage[table]{xcolor}
\usepackage{cite}
\usepackage{newtxtext,newtxmath}
\usepackage{balance}

\makeatletter
\renewcommand\footnotesize{\@setfontsize\footnotesize\@ixpt{10.5}}
\def\@ssect#1#2#3#4#5{%
  \begingroup\bf\centering
    {\interlinepenalty\@M \uppercase{#5}\par}%
  \endgroup
  \@tempskipa #3\relax
  \@xsect{\@tempskipa}}
\makeatother
\newcommand{\tabsize}{\small}

\makeatletter
\long\def\@makecaption#1#2{%
  \vskip\abovecaptionskip
  \begingroup\tabsize
  \setbox\@tempboxa\hbox{#1. #2}%
  \ifdim \wd\@tempboxa >\hsize #1. #2\par \else
    \hbox to\hsize{\hfil\box\@tempboxa\hfil}%
  \fi\endgroup}
\makeatother

\providecommand\paragraph{}
\renewcommand\paragraph[1]{\par\medskip\noindent\textbf{#1}\hspace{0.6em}\ignorespaces}

\usepackage{eso-pic}
\AddToShipoutPictureFG*{%
  \put(0,34){\makebox[\paperwidth]{\parbox{6.4in}{\centering\footnotesize
    This work has been submitted to the IEEE ICASSP for possible publication.
    Copyright may be transferred without notice, after which this version
    may no longer be accessible.}}}}

\newcommand{\ARPoolCtl}{94.3}
\newcommand{\ARPoolGap}{76.1}
\newcommand{\ARPoolRep}{18.2}

\newcommand{\ApHeadCk}{69.6}
\newcommand{\ApHeadCkCyc}{76.7}

\newcommand{\ApHeadFam}{67.6}
\newcommand{\ApHeadFamCyc}{75.4}
\newcommand{\ApHeadTolDrop}{1.7}

\newcommand{\ApMidCtl}{83.1}
\newcommand{\ApMidGap}{73.1}
\newcommand{\ApMidN}{349}
\newcommand{\ApMidRep}{10.0}

\newcommand{\BayesNewFam}{65.3}
\newcommand{\BayesNewFamHi}{93.3}
\newcommand{\BayesNewFamLo}{5.9}

\newcommand{\CapPosModels}{5}
\newcommand{\CapRatio}{0.46}

\newcommand{\CapSepPhrase}{5 of 6}

\newcommand{\CkCap}{24.7}
\newcommand{\CkCapHi}{36.5}
\newcommand{\CkCapLo}{13.7}

\newcommand{\CkErr}{10.4}
\newcommand{\CkErrHi}{13.9}
\newcommand{\CkErrLo}{5.6}

\newcommand{\CkExact}{76.7}
\newcommand{\CkExactHi}{84.4}
\newcommand{\CkExactLo}{68.4}
\newcommand{\CkExactN}{6}
\newcommand{\CkExactNPos}{6}

\newcommand{\CkRevExact}{60.0}

\newcommand{\CosyCtl}{93.3}
\newcommand{\CosyGap}{64.4}

\newcommand{\CosyN}{90}
\newcommand{\CosyRep}{28.9}

\newcommand{\DeltaMax}{1.23}

\newcommand{\DeltaRatio}{3.4}

\newcommand{\DilR}{0.59}

\newcommand{\DisN}{465}

\newcommand{\DurHiFixed}{16.7}
\newcommand{\DurHiFree}{16.7}

\newcommand{\DurHiN}{60}
\newcommand{\DurLoFixed}{76.7}
\newcommand{\DurLoFree}{43.3}

\newcommand{\DurLoN}{30}
\newcommand{\DurSplit}{12}

\newcommand{\ErrCtl}{0.0}
\newcommand{\ErrCtlHi}{0.0}

\newcommand{\ErrCtlLo}{0.0}
\newcommand{\ErrCtlN}{525}

\newcommand{\ErrRep}{-8.3}
\newcommand{\ErrRepCk}{-12.5}

\newcommand{\ExSensAll}{61.7}

\newcommand{\ExSensN}{1620}

\newcommand{\ExactCtl}{94.3}

\newcommand{\ExactRep}{18.2}
\newcommand{\ExactRepKHi}{6.8}

\newcommand{\ExactRepKLo}{40.5}

\newcommand{\ExclCapCtl}{5.6}
\newcommand{\ExclCapRep}{10.6}

\newcommand{\FamCap}{20.1}
\newcommand{\FamCapHi}{38.5}
\newcommand{\FamCapLo}{10.7}

\newcommand{\FamExact}{75.4}
\newcommand{\FamExactHi}{78.9}
\newcommand{\FamExactLo}{71.1}
\newcommand{\FamExactN}{3}
\newcommand{\FivettsCtl}{85.6}
\newcommand{\FivettsGap}{60.0}

\newcommand{\FivettsRep}{25.6}

\newcommand{\FullLadderGap}{46.8}
\newcommand{\FullLadderN}{1559}
\newcommand{\GainCtlHiLlasaEightB}{53.1}
\newcommand{\GainCtlHiLlasaOneB}{63.5}
\newcommand{\GainCtlHiLlasaThreeB}{68.0}
\newcommand{\GainCtlHiQwenOneSevenB}{45.8}
\newcommand{\GainCtlHiQwenZeroSixB}{39.6}
\newcommand{\GainCtlHiXttsTwo}{40.7}
\newcommand{\GainCtlLlasaEightB}{45.5}
\newcommand{\GainCtlLlasaOneB}{56.7}
\newcommand{\GainCtlLlasaThreeB}{58.9}
\newcommand{\GainCtlLoLlasaEightB}{37.2}
\newcommand{\GainCtlLoLlasaOneB}{50.9}
\newcommand{\GainCtlLoLlasaThreeB}{52.8}
\newcommand{\GainCtlLoQwenOneSevenB}{37.7}
\newcommand{\GainCtlLoQwenZeroSixB}{30.6}
\newcommand{\GainCtlLoXttsTwo}{29.1}
\newcommand{\GainCtlQwenOneSevenB}{41.5}
\newcommand{\GainCtlQwenZeroSixB}{34.9}
\newcommand{\GainCtlXttsTwo}{34.8}
\newcommand{\GainRepHiLlasaEightB}{20.3}
\newcommand{\GainRepHiLlasaOneB}{25.7}
\newcommand{\GainRepHiLlasaThreeB}{23.3}
\newcommand{\GainRepHiQwenOneSevenB}{37.2}
\newcommand{\GainRepHiQwenZeroSixB}{25.9}
\newcommand{\GainRepHiXttsTwo}{31.1}
\newcommand{\GainRepLlasaEightB}{15.0}
\newcommand{\GainRepLlasaOneB}{18.2}
\newcommand{\GainRepLlasaThreeB}{12.4}
\newcommand{\GainRepLoLlasaEightB}{11.0}
\newcommand{\GainRepLoLlasaOneB}{9.8}
\newcommand{\GainRepLoLlasaThreeB}{-1.8}
\newcommand{\GainRepLoQwenOneSevenB}{29.6}
\newcommand{\GainRepLoQwenZeroSixB}{16.3}
\newcommand{\GainRepLoXttsTwo}{16.9}
\newcommand{\GainRepQwenOneSevenB}{33.4}
\newcommand{\GainRepQwenZeroSixB}{20.9}
\newcommand{\GainRepXttsTwo}{24.0}

\newcommand{\GreedyGap}{72.2}

\newcommand{\GreedySampGap}{75.0}

\newcommand{\HzFormHi}{5.4}
\newcommand{\HzFormLo}{3.1}

\newcommand{\HzNModels}{4}

\newcommand{\HzRatio}{4.2}
\newcommand{\HzRevNames}{Qwen3-TTS-1.7B}

\newcommand{\IndGapHi}{76.7}
\newcommand{\IndGapLo}{65.1}

\newcommand{\IndNJudges}{four}

\newcommand{\JacPanelN}{five}
\newcommand{\JacQCounter}{24.7}

\newcommand{\JacQMax}{347.7}
\newcommand{\JacQMin}{21.0}

\newcommand{\JudgeScaleCtcMed}{1.00}

\newcommand{\JudgeScaleNCtc}{108}
\newcommand{\JudgeScaleNWhisper}{48}
\newcommand{\JudgeScaleNWhisperAll}{108}
\newcommand{\JudgeScaleNWhisperErr}{60}
\newcommand{\JudgeScaleWhisperMed}{0.25}

\newcommand{\LoopN}{540}
\newcommand{\LoopPct}{36.3}

\newcommand{\MainKept}{1558}
\newcommand{\MainNGen}{2052}

\newcommand{\MixArm}{76.4}
\newcommand{\MixHi}{89.2}
\newcommand{\MixLo}{63.2}

\newcommand{\NCapModels}{6}

\newcommand{\NInstrumented}{136}

\newcommand{\NSeeds}{three}
\newcommand{\NStimuli}{180}

\newcommand{\NllIndCtlHigherPct}{87}

\newcommand{\NllIndPairs}{54}

\newcommand{\OddK}{24}

\newcommand{\OddNKept}{1202}
\newcommand{\OddPEight}{92.9}
\newcommand{\OddPFour}{75.2}
\newcommand{\OddPOne}{8.5}
\newcommand{\OddPSix}{90.0}
\newcommand{\OddPThree}{62.3}
\newcommand{\OddPTwo}{51.2}

\newcommand{\PenFold}{4}

\newcommand{\PenNoneGap}{78.9}

\newcommand{\PenQFold}{3}
\newcommand{\PenQRepHi}{16.7}
\newcommand{\PenQRepLo}{7.8}
\newcommand{\PenRepHi}{21.1}
\newcommand{\PenRepLo}{15.6}

\newcommand{\PenZeroCtl}{100.0}
\newcommand{\PenZeroRep}{10.0}

\newcommand{\PerNCk}{four}
\newcommand{\PerNKept}{1539}
\newcommand{\PerNRules}{five}
\newcommand{\PerPEight}{93.1}
\newcommand{\PerPFour}{73.6}
\newcommand{\PerPOne}{10.4}
\newcommand{\PerPTwo}{47.9}

\newcommand{\PerRatioRecHi}{0.86}
\newcommand{\PerRatioRecLo}{0.69}
\newcommand{\PerRatioScan}{0.547}

\newcommand{\ProbeLateMed}{0.48}

\newcommand{\RasOffGap}{92.1}
\newcommand{\RasOnGap}{94.3}

\newcommand{\SatCtlLlasaEightB}{228}
\newcommand{\SatCtlLlasaOneB}{247}
\newcommand{\SatCtlLlasaThreeB}{249}
\newcommand{\SatCtlQwenOneSevenB}{133}
\newcommand{\SatCtlQwenZeroSixB}{115}
\newcommand{\SatCtlXttsTwo}{105}
\newcommand{\SatRepLlasaEightB}{188}
\newcommand{\SatRepLlasaOneB}{181}
\newcommand{\SatRepLlasaThreeB}{178}
\newcommand{\SatRepQwenOneSevenB}{107}
\newcommand{\SatRepQwenZeroSixB}{81}
\newcommand{\SatRepXttsTwo}{81}

\newcommand{\ShapePropSSE}{2.6}

\newcommand{\ShapeSatSSE}{485.3}

\newcommand{\ShufDownCk}{XTTS-v2}
\newcommand{\ShufDownLoss}{30.0}

\newcommand{\ShufNNull}{one}
\newcommand{\ShufNUp}{two}

\newcommand{\ShufRPct}{0.6}

\newcommand{\ShufUpGain}{17.8}
\newcommand{\SpecHi}{86.4}
\newcommand{\SpecLo}{34.4}
\newcommand{\SpecMed}{70.5}
\newcommand{\SpecN}{420}
\newcommand{\SpecNRev}{0}

\newcommand{\TruncPct}{51.3}

\newcommand{\VitsCtl}{58.9}
\newcommand{\VitsGap}{-6.7}

\newcommand{\VitsRep}{65.6}

\newcommand{\ExtCtlFortyEight}{46.0}
\newcommand{\ExtCtlNinetySix}{64.0}
\newcommand{\ExtCtlOneTwoEight}{66.5}
\newcommand{\ExtCtlSixtyFour}{59.5}
\newcommand{\ExtRepFortyEight}{25.0}
\newcommand{\ExtRepNinetySix}{22.0}
\newcommand{\ExtRepOneTwoEight}{17.0}
\newcommand{\ExtRepSixtyFour}{25.0}
\newcommand{\NllGptCtl}{5.34}
\newcommand{\NllGptPct}{100}
\newcommand{\NllGptRep}{3.36}
\newcommand{\NllPhiCtl}{4.89}
\newcommand{\NllPhiRep}{3.69}
\newcommand{\PhCtlLlasaEightB}{0.59}
\newcommand{\PhCtlLlasaOneB}{0.86}
\newcommand{\PhCtlLlasaThreeB}{0.83}
\newcommand{\PhCtlQwenOneSevenB}{0.89}
\newcommand{\PhCtlQwenZeroSixB}{0.95}
\newcommand{\PhRepLlasaEightB}{0.86}
\newcommand{\PhRepLlasaOneB}{1.01}
\newcommand{\PhRepLlasaThreeB}{0.99}
\newcommand{\PhRepQwenOneSevenB}{0.90}
\newcommand{\PhRepQwenZeroSixB}{0.74}

\title{Repetition, Not Length: Isolating the Counting\\
       Failure in Neural Text-to-Speech}

\name{Kirill Borodin$^{1}$, Vasilii Kudryavtsev$^{1}$, Maxim Maslov$^{2}$, Grach Mkrtchian$^{2,3}$}
\address{$^{1}$BitmanagerAI, Dubai, UAE \quad
         $^{2}$lab260, Yerevan, Armenia \quad
         $^{3}$MTUCI, Moscow, Russia \\
         kborodin.research@gmail.com}

\begin{document}
\maketitle
\def\baselinestretch{.95}\normalsize

\begin{abstract}
Text-to-speech models loop, truncate and lose count on text that repeats a
phrase many times. We show that repetition itself is what breaks them, not
the length that comes with it. Every repeated sentence in our test set is
paired with a control of matched sentence and word count in which no word
ever repeats back-to-back. Six models from three architectures render the
controls almost perfectly and fail the repeated twins: \ExactCtl{}\%
against \ExactRep{}\% exactly right at $k\ge6$. The gap survives greedy
decoding, repetition-penalty sweeps, \IndNJudges{} independent speech
recognisers and \SpecN{} analysis specifications without once reversing
sign; a held-out fourth architecture lands within a point of its predicted
gap, and one of two non-autoregressive baselines shows the same failure.
Varying the period of the text shows the failure grows smoothly with
periodicity, half of it surviving when no word is adjacent to itself.
\end{abstract}


\section{Introduction}
\label{sec:intro}

Ask a text-to-speech (TTS) model to repeat a word a dozen times and it
will frequently get the count wrong: it stops early, runs on, or locks
into a loop. The behaviour is documented as a stability
problem~\cite{song2024ellav,wang2023valle} and usually attributed to
exposure bias or missing alignment constraints, with fixes proposed at the
level of attention supervision or decoding
rules~\cite{xin2024ralle,du2024cosyvoice2,chen2024valle2}; it is the
speech analogue of neural text
degeneration~\cite{holtzman2020curious,welleck2020neural}.

Such reports confound two variables, because repeating a phrase $k$ times
makes the text both longer and repetitive. If long inputs are simply harder, nothing about repetition needs
explaining. Our design separates the two:
every repeated item is paired with a \emph{length-matched control} of the
same carrier and word count, in which the $k$ copies of one
word become non-adjacent distinct fillers, drawn from a fixed pool of
eight. A length account predicts the two curves
coincide. They do not.

We contribute: (i) a controlled manipulation isolating repetition
from length, validated on six checkpoints, two non-autoregressive
baselines and a held-out architecture; (ii) a purpose-built test set and a judge methodology for repetitive
speech; and (iii) a negative mechanistic result: the contraction
account of~\cite{viakhirev2026dispersion} does not describe any decoder
we measured.

\section{Experimental setup}
\label{sec:method}

\textbf{Test set.}
\NStimuli{} items, generated deterministically by a released script; the
texts are authored, not drawn from a corpus, since every sentence must
contain its target word a known number of times. Templates follow fixed
constraints: targets are common one- or two-syllable words
without close homophones, and carriers are plain conversational
English. Six carrier templates each fix a prefix, target and
suffix (``The dog was \emph{very} \dots{} big''); the target is repeated
$k \in \{1,2,3,4,6,8,12,16,24,32\}$ times; we call this sweep of $k$
the ladder. Sentence-repetition, tongue-twister and number-phrase item
types are built the same way. Every item at
$k\ge2$ has a length-matched control from the same template: the $k$
copies are replaced, in fixed order, by fillers from a per-template pool
of eight, so only repetition differs within a pair. The pool cycles above $k{=}8$, making those
controls period-8; a never-cycled variant quantifies the difference
(Sec.~\ref{sec:results}). Nor are the items improbable
strings (Table~\ref{tab:nll}): under two independent language models the
\emph{control} is the less probable text in most pairs, so the repeated
items are, if anything, the more natural ones. The generation script, test set, transcripts and analysis code are released.\footnote{\url{https://github.com/lab260ru/tts-counting-failure}}

\begin{table}[t]
\centering
\caption{Likelihood of the paired texts under two independent scorers:
mean NLL per token (lower = more probable) and the share of
\NllIndPairs{} pairs whose control is less probable. A panel-internal
scorer agrees (supplementary material).}
\label{tab:nll}
{\tabsize\setlength{\tabcolsep}{4pt}
\begin{tabular}{lrrr}
\toprule
Scorer & \multicolumn{2}{c}{NLL/token} & ctl.\ less probable \\
\cmidrule(lr){2-3}
 & rep. & ctl. & (\% of pairs) \\
\midrule
phi-2       & \NllPhiRep{} & \NllPhiCtl{} & \NllIndCtlHigherPct{} \\
GPT-2-large & \NllGptRep{} & \NllGptCtl{} & \NllGptPct{} \\
\bottomrule
\end{tabular}}
\end{table}

\textbf{Models.}
Six checkpoints from three architectures form the panel: Llasa at 1B/3B/8B
(Llama backbones over X-codec2 tokens~\cite{ye2025llasa}), Qwen3-TTS at
0.6B/1.7B~\cite{hu2026qwen3tts}, and XTTS-v2~\cite{casanova2024xtts}.
Outside the panel: VITS~\cite{kim2021vits} and F5-TTS~\cite{chen2024f5tts}
as non-autoregressive baselines, CosyVoice~2~\cite{du2024cosyvoice2} as a
fourth architecture held out until the
analysis was frozen, and penalty, greedy and sampling ablations of panel
members (Tables~\ref{tab:models} and~\ref{tab:vary}). Every system renders every item at
\NSeeds{} seeds under each system's shipped decoding defaults. The defaults make
Qwen3-TTS the most stochastic family, so its larger checkpoint's
near-zero count-error gap (Sec.~\ref{sec:results}) is not conservative
decoding at work.

\textbf{Judge.}
The judge, the speech recogniser whose transcript we count, is a CTC
model ({\ttfamily wav2vec2-\allowbreak large-\allowbreak
960h-\allowbreak lv60-\allowbreak self}; greedy best-path, no language
model). The natural choice, Whisper large-v3~\cite{radford2023whisper}, is
unsuitable here: its decoder is autoregressive with a language-model prior
and hallucinates on exactly the audio under
study~\cite{viakhirev2026dispersion}. On concatenated utterances whose true count is known by construction,
Whisper recovers a median
\JudgeScaleWhisperMed{} of the count on the \JudgeScaleNWhisper{} trials
where it returns anything, and nothing at all in the other
\JudgeScaleNWhisperErr{} of \JudgeScaleNWhisperAll{}; the CTC judge
recovers \JudgeScaleCtcMed{} ($n{=}\JudgeScaleNCtc{}$) on the same audio.
The full audit and the \IndNJudges{}-recogniser replication are in the
supplementary material.

\textbf{Metrics.}
$\hat c$ is the number of renditions of the target found in the transcript,
counted in order and non-overlapping, skipping unrecognised words
(stopping at the first miss
would void every later filler). Three quantities are reported. \emph{Relative count error}:
$(\hat c - k)/k$, negative for undercounts. \emph{Exactly right}: the share
of generations with $\hat c = k$. \emph{Gap}: control minus repeated
exactly-right rate, in percentage points; seen from the repeated side we
also call it the deficit. Headline analyses cover the word-repetition ladder and its controls
(other item types: supplementary material). Exclusion rules blind to the
count (a template the judge cannot transcribe, our token budget,
degenerate audio) keep \MainKept{} of \MainNGen{} generations; the accounting is in the
supplementary material and Table~\ref{tab:vary} shows the gap with every
rule off. Individual analyses run on slightly
different populations, so each number carries its own $n$.

\textbf{Uncertainty.}
Every bracketed range $[\cdot,\cdot]$ in this paper is a 95\% confidence
interval: Wilson for single rates; for panel summaries, a nonparametric
bootstrap over checkpoints, the unit our claims generalise over
(generations cluster within one). Every headline estimate is computed
four ways (per checkpoint, per family, mixed-effects and hierarchical Bayes;
Table~\ref{tab:ci}) and under
\SpecN{} specifications, so no conclusion rests on one analysis choice.

\section{Results}
\label{sec:results}

\begin{figure}[t]
\centering
\includegraphics[width=\columnwidth]{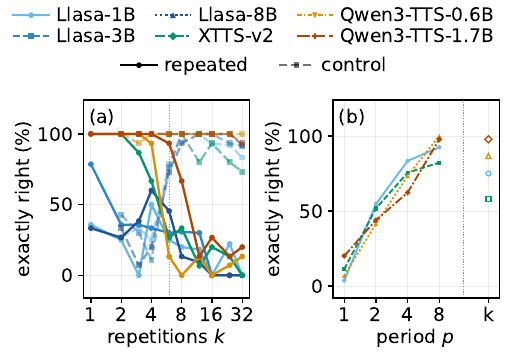}
\caption{The repeated--control dissociation, in the paper's headline
metric. (a) Exactly-right renderings against $k$: from $k{=}6$, length-matched
controls (dashed) stay near-perfect while repeated text (solid) collapses;
below $k{=}6$ transcription noise affects both conditions. (b) On the four checkpoints measured, at fixed
carrier, word count and requested count, exact renderings rise smoothly with period $p$; hollow
markers: the all-distinct $p{=}k$ items.}
\label{fig:main}
\end{figure}

\subsection{The limit tracks repetition, not length}

Figure~\ref{fig:main}(a) is the central test. Each control
carries its twin's carrier and word count, so any account in which
length is what hurts (attention span, exposure bias, duration budgets)
predicts the two curves coincide.
They do not: from $k{=}6$ the controls hold at \ExactCtl{}\% exactly
right while repeated items fall from \ExactRepKLo{}\% at $k{=}6$ to
\ExactRepKHi{}\% at $k{=}32$, \ExactRep{}\% pooled
(Table~\ref{tab:models}). Below $k{=}6$ the two curves do not separate.

The miscounts are large and biased toward undercounting. Repeated items
run a median count error of $\ErrRepCk{}$\% per checkpoint against
\ErrCtl{}\% $[\ErrCtlLo{},\ErrCtlHi{}]$ for controls
($n{=}\ErrCtlN{}$), at every $k$ up to $32$, in five of six checkpoints;
the sixth ties at zero on this metric yet still loses \CkRevExact{}
points of exact rate. Before exclusions, at $k{\ge}8$ ($n{=}\LoopN{}$),
\TruncPct{}\% truncate or undercount and \LoopPct{}\% loop or overcount.

Table~\ref{tab:ci} gives the headline estimates under every clustering
treatment. The exact-rate gap is positive in all \CkExactNPos{} of \CkExactN{}
checkpoints and all \FamExactN{} families, and all six interval
constructions of the mixed-effects model exclude zero.

The statistical model also predicts out of sample: partial pooling of
the three architectures puts the gap for an unseen architecture at
\BayesNewFam{} $[\BayesNewFamLo{},\BayesNewFamHi{}]$ points; the widest
prior widens that interval to include zero, so we tested the prediction. CosyVoice~2, the design most likely to be immune thanks to
its alignment supervision, its numbers predicted before it generated a
waveform, lands within a point of the prediction, at \CosyGap{}
(Tables~\ref{tab:ci} and~\ref{tab:models}).

The $k{\ge}6$ threshold was chosen after seeing where the curves diverge, so
we also quote the unconditioned whole-ladder gap (Table~\ref{tab:ci}) and
vary the threshold along with every other analysis choice: across \SpecN{}
specifications over six axes (exclusions, threshold, control variant,
seeds, aggregation, outcome definition) the gap spans
\SpecLo{}--\SpecHi{} points, median \SpecMed{}, and \SpecNRev{} of \SpecN{}
reverse sign (Table~\ref{tab:vary}).

Above $k{=}8$ the cycled control is itself period-8, and the headline
gaps above use that cycled control. Using never-cycled fillers throughout
shrinks the per-family $k{\ge}6$ gap from \ApHeadFamCyc{} to
\ApHeadFam{} points and the per-checkpoint one from \ApHeadCkCyc{} to
\ApHeadCk{}. Most of the difference is
vocabulary difficulty, not periodicity: \ApHeadTolDrop{} points survive one
unit of counting tolerance, and none of it shows on median error.
Re-generated at matched $k$ (12--32) with fully
aperiodic controls, the rates are \ApMidRep{}\% against \ApMidCtl{}\%
($n{=}\ApMidN{}$): \ApMidGap{} points with no cycling anywhere.

\begin{table}[t]
\centering
\caption{Headline estimates with 95\% intervals, in exactly-right points
(the count-error row: points of median relative error). The last two rows
are the pre-registered out-of-panel prediction and its test.}
\label{tab:ci}
{\tabsize\setlength{\tabcolsep}{3.5pt}
\begin{tabular}{lrr}
\toprule
Quantity & est. & 95\% interval \\
\midrule
Exact-rate gap, per checkpoint ($n{=}\CkExactN{}$) & \CkExact{} & $[\CkExactLo{},\,\CkExactHi{}]$ \\
Exact-rate gap, per family ($n{=}\FamExactN{}$)    & \FamExact{} & $[\FamExactLo{},\,\FamExactHi{}]$ \\
Exact-rate gap, mixed effects (widest of six)      & \MixArm{}  & $[\MixLo{},\,\MixHi{}]$ \\
Count-error gap, per checkpoint                    & \CkErr{}   & $[\CkErrLo{},\,\CkErrHi{}]$ \\
Exact-rate gap, whole ladder ($n{=}\FullLadderN{}$) & \FullLadderGap{} & --- \\
\midrule
Predicted gap, unseen architecture                 & \BayesNewFam{} & $[\BayesNewFamLo{},\,\BayesNewFamHi{}]$ \\
Observed, CosyVoice~2~\cite{du2024cosyvoice2} ($n{=}\CosyN{}$ per arm) & \CosyGap{} & --- \\
\bottomrule
\end{tabular}}
\end{table}

\begin{table}[t]
\centering
\caption{Every system tested, $k\ge6$, CTC judge: median $(\hat c-k)/k$
and exactly-right rate, repeated (rep.) against control (ctl.). Systems in
the lower block are outside the panel; their count errors are in the
supplementary material.}
\label{tab:models}
{\tabsize\setlength{\tabcolsep}{2.5pt}
\begin{tabular}{llrrrr}
\toprule
Model & size & \multicolumn{2}{c}{count err.} & \multicolumn{2}{c}{exact (\%)} \\
\cmidrule(lr){3-4}\cmidrule(lr){5-6}
 & & rep. & ctl. & rep. & ctl. \\
\midrule
Llasa-1B~\cite{ye2025llasa} & 1B & $-12.5$ & 0.0 & 13.3 & 91.5 \\
Llasa-3B~\cite{ye2025llasa} & 3B & $-8.3$ & 0.0 & 16.9 & 94.1 \\
Llasa-8B~\cite{ye2025llasa} & 8B & $-16.7$ & 0.0 & 11.8 & 93.2 \\
XTTS-v2~\cite{casanova2024xtts} & 0.4B & $-12.5$ & 0.0 & 16.7 & 87.8 \\
Qwen3-TTS-0.6B~\cite{hu2026qwen3tts} & 0.6B & $-12.5$ & 0.0 & 7.9 & 100.0 \\
Qwen3-TTS-1.7B~\cite{hu2026qwen3tts} & 1.7B & 0.0 & 0.0 & 38.9 & 98.9 \\
\midrule
Panel, pooled & 6 ckpts. & $\ErrRep{}$ & \ErrCtl{} & \ARPoolRep{} & \ARPoolCtl{} \\
CosyVoice~2~\cite{du2024cosyvoice2} & held-out & --- & --- & \CosyRep{} & \CosyCtl{} \\
F5-TTS~\cite{chen2024f5tts} & non-AR & --- & --- & \FivettsRep{} & \FivettsCtl{} \\
VITS~\cite{kim2021vits} & non-AR & --- & --- & \VitsRep{} & \VitsCtl{} \\
\bottomrule
\end{tabular}
}
\end{table}

\textbf{Extension to $k\le128$ (exploratory).}
Within $k\le32$ the deficit is proportional in $k$, not saturating
(residual \ShapePropSSE{} against \ShapeSatSSE{} for a saturating fit), so
no collapse point lies inside the main ladder. Extending \HzNModels{}
checkpoints to $k\le128$ shows the two conditions diverging in absolute terms
(Table~\ref{tab:ext}): the median rendered count of repeated items
\emph{falls} as the requested count rises, while controls keep climbing.
The fitted saturation scale drops \HzRatio{}-fold, but the factor
depends on the fitted form (\HzFormLo{}--\HzFormHi{} across three
saturating forms) and is not unanimous (in \HzRevNames{} it moves the
other way), so this cannot tell a hard limit from ordinary saturation.

\begin{table}[t]
\centering
\caption{The $k\le128$ extension, \HzNModels{} checkpoints: median
\emph{rendered} count against the requested count. Repeated renderings
regress while controls advance.}
\label{tab:ext}
{\tabsize\setlength{\tabcolsep}{6pt}
\begin{tabular}{lrr}
\toprule
Requested count & repeated & control \\
\midrule
$k{=}48$  & \ExtRepFortyEight{} & \ExtCtlFortyEight{} \\
$k{=}64$  & \ExtRepSixtyFour{} & \ExtCtlSixtyFour{} \\
$k{=}96$  & \ExtRepNinetySix{} & \ExtCtlNinetySix{} \\
$k{=}128$ & \ExtRepOneTwoEight{} & \ExtCtlOneTwoEight{} \\
\bottomrule
\end{tabular}}
\end{table}

\subsection{The deficit is graded in the period}

\begin{table}[t]
\centering
\caption{The period ladder: exactly-right rate (\%) against the period $p$
of the text at fixed carrier, word count and requested count, pooled over
the \PerNCk{} checkpoints measured. The second row re-measures at
$k{=}\OddK{}$ with the odd and non-power steps~added.}
\label{tab:period}
{\tabsize\setlength{\tabcolsep}{4.5pt}
\begin{tabular}{lrrrrrr}
\toprule
 & $p{=}1$ & $p{=}2$ & $p{=}3$ & $p{=}4$ & $p{=}6$ & $p{=}8$ \\
\midrule
Ladder scan ($n{=}\PerNKept{}$) & \PerPOne{} & \PerPTwo{} & --- & \PerPFour{} & --- & \PerPEight{} \\
At $k{=}\OddK{}$ ($n{=}\OddNKept{}$) & \OddPOne{} & \OddPTwo{} & \OddPThree{} & \OddPFour{} & \OddPSix{} & \OddPEight{} \\
\bottomrule
\end{tabular}}
\end{table}

So far, repetition and periodicity vary together; a second ladder holds
carrier, word count and requested count fixed and varies only the period
$p$. The deficit is graded, not a step: exact rates rise
\PerPOne{}, \PerPTwo{}, \PerPFour{}, \PerPEight{}\% at $p{=}1,2,4,8$,
monotone in all \PerNCk{} checkpoints measured (Table~\ref{tab:period};
Fig.~\ref{fig:main}b).

The decisive step is $p{=}2$: no token is ever adjacent to itself, yet
half the deficit remains (\PerRatioScan{} on the strictest of
\PerNRules{} scoring rules, \PerRatioRecLo{}--\PerRatioRecHi{} on the
others). Nor is it a
power-of-two artifact: at $k{=}\OddK{}$ the odd step $p{=}3$
(\OddPThree{}\%) and non-power $p{=}6$ (\OddPSix{}\%) land between
their neighbours, as pre-registered. A comma,
full stop or ``and'' between repetitions recovers none of it
($n{=}\DisN{}$): the copies are segmented, just miscounted.

Whether the \emph{ordering} of a period-$p$ text or its restricted
\emph{vocabulary} carries the effect is unresolved. Shuffling each item's own words recovers almost none of the gap pooled
(\ShufRPct{}\%), but that pooled number averages over a split:
\ShufNUp{} checkpoints gain \ShufUpGain{} points, \ShufNNull{} is null,
and \ShufDownCk{} loses \ShufDownLoss{}. That split is why the title claims
repetition rather than periodicity.

\subsection{Not the decoding rule or our analysis choices}

\begin{table}[t]
\centering
\caption{The gap under decoding and analysis variations. RAS is
repetition-aware sampling~\cite{chen2024valle2}.}
\label{tab:vary}
{\tabsize\setlength{\tabcolsep}{3pt}
\begin{tabular}{lr}
\toprule
 & gap (pts) \\
\midrule
Panel, $k\ge6$                    & \ARPoolGap{} \\
Exclusions off ($n{=}\ExSensN{}$) & \ExSensAll{} \\
No penalty (Llasa $\times$3)      & \PenNoneGap{} \\
Greedy (Qwen3-TTS-0.6B)                   & \GreedyGap{} \\
Sampled (Qwen3-TTS-0.6B)                  & \GreedySampGap{} \\
RAS on (Qwen3-TTS-0.6B)                   & \RasOnGap{} \\
RAS off (Qwen3-TTS-0.6B)                  & \RasOffGap{} \\
Across \IndNJudges{} recognisers (range) & \IndGapLo{}--\IndGapHi{} \\
Across \SpecN{} specifications (range)  & \SpecLo{}--\SpecHi{} \\
\bottomrule
\end{tabular}}
\end{table}

Table~\ref{tab:vary} varies the decoding rule and the analysis. The families ship penalties differing by an order of magnitude; swept
over \PenFold{}-fold (XTTS-v2) and \PenQFold{}-fold (Qwen3-TTS-0.6B)
ranges, the gap stays put (\PenRepLo{}--\PenRepHi{}\% and
\PenQRepLo{}--\PenQRepHi{}\% exact). The three Llasa checkpoints apply no
penalty at all and show it at full size; with the penalty disabled
entirely, Qwen3-TTS-0.6B renders \PenZeroRep{}\% exactly against
\PenZeroCtl{}\%.

Greedy decoding and repetition-aware sampling leave the gap in place. Nor do our exclusion rules create the gap: with every rule off it is
\ExSensAll{} points. The one rule that falls unevenly, the token budget
(hit by \ExclCapRep{}\% of repeated against \ExclCapCtl{}\% of control
generations), can
only cut counts short; keeping those rows would widen the gap, so
excluding them is conservative. Re-scored
with \IndNJudges{} independent recognisers it stays at
\IndGapLo{}--\IndGapHi{} points, positive everywhere; the per-judge
analysis is in the supplementary material.

\subsection{Non-autoregressive baselines}

VITS shows no dissociation on the same strings through the same judge:
$\VitsGap{}$ points against the panel's \ARPoolGap{}
(Table~\ref{tab:models}). F5-TTS, flow-matching rather than
autoregressive, shows a \FivettsGap{}-point
gap, so the failure is not specific to autoregressive decoding.

What plausibly separates the two is where the count must live: VITS
predicts a duration per token, F5-TTS one total. Handed the oracle total duration, F5-TTS improves markedly below
$k{=}\DurSplit{}$ and not at all at or above it (Table~\ref{tab:dur}):
given the right length, the model still cannot place thirty-two
repetitions inside it.

\begin{table}[t]
\centering
\caption{F5-TTS handed the oracle total duration: exactly-right
rate~(\%).}
\label{tab:dur}
{\tabsize\setlength{\tabcolsep}{5pt}
\begin{tabular}{lrr}
\toprule
 & free & oracle duration \\
\midrule
Below $k{=}\DurSplit{}$ ($n{=}\DurLoN{}$)      & \DurLoFree{} & \DurLoFixed{} \\
At or above $k{=}\DurSplit{}$ ($n{=}\DurHiN{}$) & \DurHiFree{} & \DurHiFixed{} \\
\bottomrule
\end{tabular}}
\end{table}

\subsection{Repetition slows the growth of the state}
\label{sec:cap}

The failure is also visible in the decoder's state space. On a
\NInstrumented{}-item subset we record per-step hidden
states at deep probe layers and compute the trajectory's effective rank
$N_{\mathrm{eff}} = \exp(-\sum_i p_i \log p_i)$, $p_i$ the normalised
singular values: the number of distinguishable states visited.
The statistic is the \emph{capacity gain} $dN_{\mathrm{eff}}/d\log k$,
fitted separately on repeated items and on their length-matched
controls.

Repeated text grows new states at \CapRatio{} of its control's rate, median
across the panel, with the two intervals disjoint in \CapSepPhrase{}
checkpoints (Table~\ref{tab:cap}). The gain stays bounded away from zero in \CapPosModels{} of
\NCapModels{}, so this is a slowing, not a halt. At
checkpoint level the gap in gains is \CkCap{} $[\CkCapLo{},\CkCapHi{}]$
states per unit $\log k$; by family, \FamCap{} $[\FamCapLo{},\FamCapHi{}]$. The ceiling columns of Table~\ref{tab:cap} agree.

\begin{table}[t]
\centering
\caption{State capacity per checkpoint: growth of distinguishable states
$dN_{\mathrm{eff}}/d\log k$ with 95\% intervals, and the saturating-fit
ceiling of $N_{\mathrm{eff}}$, repeated against control.}
\label{tab:cap}
{\tabsize\setlength{\tabcolsep}{1.4pt}
\begin{tabular}{lrrrr}
\toprule
Model & \multicolumn{2}{c}{capacity gain} & \multicolumn{2}{c}{ceiling} \\
\cmidrule(lr){2-3}\cmidrule(lr){4-5}
 & rep. & ctl. & rep. & ctl. \\
\midrule
Llasa-1B~\cite{ye2025llasa} & \GainRepLlasaOneB{} $[\GainRepLoLlasaOneB{}{,}\GainRepHiLlasaOneB{}]$ & \GainCtlLlasaOneB{} $[\GainCtlLoLlasaOneB{}{,}\GainCtlHiLlasaOneB{}]$ & \SatRepLlasaOneB{} & \SatCtlLlasaOneB{} \\
Llasa-3B~\cite{ye2025llasa} & \GainRepLlasaThreeB{} $[\GainRepLoLlasaThreeB{}{,}\GainRepHiLlasaThreeB{}]$ & \GainCtlLlasaThreeB{} $[\GainCtlLoLlasaThreeB{}{,}\GainCtlHiLlasaThreeB{}]$ & \SatRepLlasaThreeB{} & \SatCtlLlasaThreeB{} \\
Llasa-8B~\cite{ye2025llasa} & \GainRepLlasaEightB{} $[\GainRepLoLlasaEightB{}{,}\GainRepHiLlasaEightB{}]$ & \GainCtlLlasaEightB{} $[\GainCtlLoLlasaEightB{}{,}\GainCtlHiLlasaEightB{}]$ & \SatRepLlasaEightB{} & \SatCtlLlasaEightB{} \\
XTTS-v2~\cite{casanova2024xtts} & \GainRepXttsTwo{} $[\GainRepLoXttsTwo{}{,}\GainRepHiXttsTwo{}]$ & \GainCtlXttsTwo{} $[\GainCtlLoXttsTwo{}{,}\GainCtlHiXttsTwo{}]$ & \SatRepXttsTwo{} & \SatCtlXttsTwo{} \\
Qwen3-TTS-0.6B~\cite{hu2026qwen3tts} & \GainRepQwenZeroSixB{} $[\GainRepLoQwenZeroSixB{}{,}\GainRepHiQwenZeroSixB{}]$ & \GainCtlQwenZeroSixB{} $[\GainCtlLoQwenZeroSixB{}{,}\GainCtlHiQwenZeroSixB{}]$ & \SatRepQwenZeroSixB{} & \SatCtlQwenZeroSixB{} \\
Qwen3-TTS-1.7B~\cite{hu2026qwen3tts} & \GainRepQwenOneSevenB{} $[\GainRepLoQwenOneSevenB{}{,}\GainRepHiQwenOneSevenB{}]$ & \GainCtlQwenOneSevenB{} $[\GainCtlLoQwenOneSevenB{}{,}\GainCtlHiQwenOneSevenB{}]$ & \SatRepQwenOneSevenB{} & \SatCtlQwenOneSevenB{} \\
\bottomrule
\end{tabular}}
\end{table}

\section{Discussion}
\label{sec:discussion}

\textbf{A mechanism tried, and rejected.}
The contraction account of decoder
hallucination~\cite{viakhirev2026dispersion} attributes looping to
attractor dynamics: repeated conditioning drives the decoder state toward a
fixed point, and once states have contracted together no readout can
recover the count. It motivated this study; it fails here.

We estimated the contraction factor $q$: the largest singular value of
the map from the decoder state at one repetition boundary to the next.
$q<1$ means states drift together; $q>1$, that they separate. Across \JacPanelN{} checkpoints $q$ runs \JacQMin{}--\JacQMax{}:
expansive in every item, on repeated and control text alike. It grows with
scale inside Llasa, and the one checkpoint with zero median count error
sits at $q{=}\JacQCounter{}$. Nor is attention near-uniform over the $k$ copies (logit spread
\DeltaMax{} nats, a \DeltaRatio{}-fold ratio), and flatter attention
predicts \emph{better} counting ($r{=}\DilR{}$), the opposite of the
account's prediction.

\textbf{What fails is not storage.}
The requested count stays decodable from late hidden states while the
rendering fails (ridge probe, median $R^2{=}\ProbeLateMed{}$), as in text
models~\cite{venkatesh2026repeated}. Past the ladder ($k{=}48$--$128$) that decodability thins
(Table~\ref{tab:probe}): one checkpoint fails to beat a constant
predictor, a second ties it (\PhRepLlasaThreeB{}), and on the two Qwen
checkpoints, where the probe still wins, a probe given only the audio
length does as well, so the surviving signal need not encode the count. Deciding between state and readout needs an intervention at
fixed state; our rank-1 patches fail their own positive control, so the
question stays open.

\begin{table}[t]
\centering
\caption{Probe MAE past the ladder ($k{=}48$--$128$) as a fraction of a
constant predictor's; below 1 the count is still decodable. The control
column runs the same probe on control items.}
\label{tab:probe}
{\tabsize\setlength{\tabcolsep}{6pt}
\begin{tabular}{lrr}
\toprule
Checkpoint & repeated & control \\
\midrule
Llasa-1B~\cite{ye2025llasa} & \PhRepLlasaOneB{} & \PhCtlLlasaOneB{} \\
Llasa-3B~\cite{ye2025llasa} & \PhRepLlasaThreeB{} & \PhCtlLlasaThreeB{} \\
Llasa-8B~\cite{ye2025llasa} & \PhRepLlasaEightB{} & \PhCtlLlasaEightB{} \\
Qwen3-TTS-0.6B~\cite{hu2026qwen3tts} & \PhRepQwenZeroSixB{} & \PhCtlQwenZeroSixB{} \\
Qwen3-TTS-1.7B~\cite{hu2026qwen3tts} & \PhRepQwenOneSevenB{} & \PhCtlQwenOneSevenB{} \\
\bottomrule
\end{tabular}}
\end{table}

\textbf{Related work.}
Closest is representational collapse~\cite{barbero2024glasses};
Sec.~\ref{sec:cap} measures it under a length-matched control and finds a
slowing, not a fixed point. Counting and state-tracking
studies~\cite{yehudai2024when,salzer2025counting,zhang2024counting,%
hahn2020theoretical,strobl2024formal,li2025statetracking} delimit what transformers can count, but those limits are model
properties, invariant to the period of the input; our effect moves with
it.

\textbf{Limits.}
Repetition is not separated from training-set
frequency, though Table~\ref{tab:nll} points the wrong way for the
obvious form of that confound. The judge is
validated on audio, not listeners; the panel, six checkpoints in three
families, is as wide as compute allowed.

\section{Conclusion}
\label{sec:conclusion}

With length held fixed, repetition alone breaks every autoregressive
text-to-speech system we tested: six checkpoints from three architectures
render controls exactly in \ExactCtl{}\% and
repeated text in \ExactRep{}\%: a \ARPoolGap{}-point
gap, per checkpoint \CkExact{} $[\CkExactLo{},\CkExactHi{}]$, that
nothing we varied reverses (\SpecLo{}--\SpecHi{} across \SpecN{}
specifications). The gap is graded in the text's period, half
remaining with no adjacent repeats. It extends to a held-out architecture
(\CosyGap{} points, as predicted) and one of two parallel decoders: not
autoregression itself. Repeated text grows distinguishable states at
\CapRatio{} of the control's rate, and the attractor mechanism fails its
own measurement ($q$ never below~1). Why the readout fails is the
open question.

\section*{Compliance with Ethical Standards}
This is a computational study of publicly released text-to-speech
checkpoints, evaluated on synthetic text written by the authors. It
involved no human subjects, no animal subjects and no personal data, and
no ethical approval was required. The authors declare no competing
financial interests.

\section*{Acknowledgements}
The authors used Claude Fable~5 (Anthropic) only to polish the language
of this paper. All ideas, experimental design, experiments and analysis
are the authors' own; the authors reviewed all text and take full
responsibility for it. The authors declare no conflicts of interest.

\let\oldthebibliography\thebibliography
\renewcommand{\thebibliography}[1]{%
  \oldthebibliography{#1}%
  \setlength{\itemsep}{0pt}%
  \setlength{\parsep}{0pt}%
  \setlength{\parskip}{0pt}}
\balance   
{\tabsize
\bibliographystyle{IEEEbib}
\bibliography{refs}}

\clearpage
\nobalance
\onecolumn
\twocolumn[{\centering\LARGE\bfseries Supplementary Material\par\bigskip}]
\appendix
\setcounter{table}{0}
\renewcommand{\thetable}{S\arabic{table}}
\renewcommand{\theHtable}{S\arabic{table}}

\setcounter{topnumber}{4}
\setcounter{dbltopnumber}{4}
\setcounter{totalnumber}{8}
\renewcommand{\topfraction}{.95}
\renewcommand{\dbltopfraction}{.95}
\renewcommand{\textfraction}{.03}
\renewcommand{\floatpagefraction}{.75}
\renewcommand{\dblfloatpagefraction}{.75}
\makeatletter
\setlength{\@fptop}{0pt}
\setlength{\@fpbot}{0pt plus 1fil}
\setlength{\@dblfptop}{0pt}
\setlength{\@dblfpbot}{0pt plus 1fil}
\newcommand{\inlinecaption}[1]{\refstepcounter{table}\@makecaption{\fnum@table}{#1}\vskip\belowcaptionskip}
\makeatother

\section{The test set in full}
\label{sec:testset}

The test set contains $180$ items in five families
(Table~\ref{tab:families}); scoring counts occurrences of a target unit
against an expected count, not string matches against a reference
transcript.

\begin{table*}[tp]
\centering
\caption{Item families.}
\label{tab:families}
\begin{tabular}{lrl}
\toprule
Family & Count & What it tests \\
\midrule
word repetition & 60 & a single word repeated $k$ times in a fixed carrier \\
matched control & 54 & the same carrier, $k$ \emph{distinct} filler words (no repetition) \\
sentence repetition & 32 & a whole short sentence repeated $k$ times \\
tongue twisters & 24 & a twister sentence (with internal repetition) repeated $k$ times \\
numbers & 10 & English number phrases with intrinsic digit-word repetition \\
\midrule
Total & 180 & \\
\bottomrule
\end{tabular}
\end{table*}

\paragraph{The $k$-ladder.} Word repetition and controls run
$k \in \{1,2,3,4,$ $6,8,12,16,$ $24,32\}$, bracketing the expected collapse
point $k^\ast \in 4$--$16$; sentence repetition stops at $k=16$, twisters
at $k=4$ (Table~\ref{tab:twisters}); numbers are fixed strings, no
ladder (Table~\ref{tab:numbers}).

\paragraph{Word-repetition carriers.} Table~\ref{tab:templates} reproduces
all six carrier templates verbatim. The six targets are common, short,
phonetically distinct from every filler, and homophone-free, limiting
judge mis-transcription noise; the judge audit (Section~\ref{sec:judge})
later located the transcription risk in the judge's autoregressive
decoder, not the word list.
\begin{table*}[tp]
\centering
\caption{Word-repetition carriers: fixed prefix, target repeated $k$ times,
fixed suffix; the filler pool builds the matched control at the same $k$.}
\label{tab:templates}
\small
\begin{tabular}{@{}lp{3.2cm}p{1.4cm}p{4.4cm}p{6.6cm}@{}}
\toprule
ID & Prefix & Target & Suffix & Filler pool (for the control) \\
\midrule
t1 & The dog was & very & big and it ran across the field. & quite, really, truly, fairly, rather, somewhat, extremely, notably \\
t2 & She said & no & to the offer and then left the room. & yes, maybe, sure, fine, okay, well, right, hmm \\
t3 & He walked & far & into the forest before turning back. & deep, fast, long, wide, high, low, near, past \\
t4 & The light was & blue & before the storm arrived. & green, bright, pale, dim, warm, cold, sharp, soft \\
t5 & The runner moved & quick & along the narrow path. & swift, smooth, steady, light, sharp, loose, tight, clean \\
t6 & They were & really & tired after the long journey. & truly, quite, very, rather, deeply, clearly, plainly, surely \\
\bottomrule
\end{tabular}
\end{table*}

\paragraph{Number phrases.} Table~\ref{tab:numbers} reproduces all ten
phrases verbatim. One digit-word recurs at several magnitudes, so correct
rendering needs a place-value counter, not adjacent-token runs.
\begin{table*}[tp]
\centering
\caption{Number-phrase items, with the target digit-word and its true count.}
\label{tab:numbers}
\small
\begin{tabular}{@{}lp{12.3cm}ll@{}}
\toprule
ID & Phrase & Target & Count \\
\midrule
n01 & six hundred sixty six thousand six hundred sixty six & six & 6 \\
n02 & seven hundred seventy seven thousand seven hundred seventy seven & seven & 6 \\
n03 & nine hundred ninety nine thousand nine hundred ninety nine & nine & 6 \\
n04 & three hundred thirty three thousand three hundred thirty three & three & 6 \\
n05 & six hundred sixty six million six hundred sixty six thousand six hundred sixty six & six & 9 \\
n06 & eight hundred eighty eight thousand eight hundred eighty eight & eight & 6 \\
n07 & five hundred fifty five & five & 3 \\
n08 & four hundred forty four & four & 3 \\
n09 & two hundred twenty two million two hundred twenty two thousand two hundred twenty two & two & 9 \\
n10 & one hundred eleven thousand one hundred eleven & one & 5 \\
\bottomrule
\end{tabular}
\end{table*}

\paragraph{Sentence repetition and tongue twisters.} Sentence items
(Table~\ref{tab:sentences}) score one content word at expected count $k$;
twister items expect per-copy occurrences $\times\,k$, the w6 and w7
targets (per-copy $0$) probing phonological stress only, outside the
counting metric.
\begin{table*}[tp]
\centering
\caption{Sentence-repetition items ($k \in \{1,2,3,4,6,8,12,16\}$,
expected count $k$) and tongue-twister items ($k\in\{1,2,4\}$, expected
count per-copy occurrences of the target $\times\,k$).}
\label{tab:sentences}\label{tab:twisters}
\small
\begin{tabular}{@{}lp{11.5cm}ll@{}}
\toprule
ID & Sentence & Target & Per-copy \\
\midrule
s1 & The bell rang twice. & bell & --- \\
s2 & He counted the stones. & stones & --- \\
s3 & Rain fell on the roof. & rain & --- \\
s4 & The door stayed open. & door & --- \\
\midrule
w1 & She sells sea shells by the sea shore. & sea & 2 \\
w2 & Peter Piper picked a peck of pickled peppers. & peck & 1 \\
w3 & How much wood would a woodchuck chuck. & chuck & 1 \\
w4 & Red lorry yellow lorry red lorry yellow lorry. & lorry & 4 \\
w5 & The sixth sick sheikh's sixth sheep is sick. & sixth & 2 \\
w6 & Six slick slim sycamore saplings. & s (alliteration only) & 0 \\
w7 & Fresh French fried fly fritters. & fr (alliteration only) & 0 \\
w8 & Truly rural truly rural truly rural. & rural & 3 \\
\bottomrule
\end{tabular}
\end{table*}

\paragraph{Matched controls.} Same prefix, suffix, and word count, the
$k$ target copies replaced by $k$ distinct fillers cycled from the pool,
the target never appearing in its own control. Only periodicity is
removed: an item-control difference cannot reflect sequence length,
generation duration, or text amount, the basis of the main text's
dissociation test.

\paragraph{Repetition-boundary lists.} Word- and sentence-repetition
items list their $k$ target copies in order, controls their $k$ fillers;
the boundary-location step (Section~\ref{sec:instrumentation}) searches
one explicit, model-agnostic list, left-to-right and non-rewinding in
both arms, so fitted decay rates compare directly.

\paragraph{The instrumented subset.} Teacher-forced hidden-state and
attention capture (Section~\ref{sec:instrumentation}) covers word
repetition, sentence repetition, and controls at $k\ge2$, $136$ items;
twisters and numbers are behavioural-only.

\section{Scoring}
\label{sec:scoring}

\noindent\textbf{Two flags.} As built, Count A is Whisper
large-v3~\cite{radford2023whisper}; the judge audit
(Section~\ref{sec:judge}) showed its autoregressive decoder
systematically undercounts correct periodic speech, so the main text uses
a CTC judge scored by relative count error, and Whisper-derived numbers
are flagged where they recur. A judge swap changes only the transcript
source: Count B is recalibrated per judge, and every
transcript-independent piece survives.

\paragraph{Three metrics.} \textbf{Count A} counts exact whole-token
target matches in the normalised transcript (lower-cased, punctuation
stripped, numerals expanded to number words). \textbf{Count B} inverts a
per-model, per-family, per-template linear fit of duration on $k$,
calibrated only on a trusted low-$k$ regime ($k\le3$, Count A exact,
duration $>0.2$\,s) fixed before seeing high-$k$ behaviour. An item is
\textbf{correct} only if Count A exactly equals the expected count
\emph{and} the duration ratio against the fit lies in $[0.70, 1.45]$.

\paragraph{Exclusions.} Audio that is empty or degenerate (duration
$<0.25$\,s, RMS $<10^{-3}$, spectral flatness $>0.35$, or an empty
transcript) carries no count and is reported as its own rate, checked
before any transcript-based label. At
$|\text{Count A} - \text{Count B}| > \max(1,\,
0.15\cdot\text{expected count})$ the item is a flagged disagreement,
never a synthesised number.

\section{The judge is part of the experiment}\label{sec:judge}
On ground truth known by construction, the natural instrument, Whisper large-v3, scores $0.25$ of the true
repetition count; the CTC recogniser that replaced it scores $1.00$. Whisper's decoder is autoregressive with a
language-model prior: it discounts a further verbatim repeat and closes the sentence on repeated audio, the failure
mode under study, in the instrument that must certify it. Every earlier Whisper-judged number is flagged where it
recurs.

\paragraph{Head-to-head on synthetic ground truth.}
Six verified $k{=}1$ atoms per arm, a target word rendered exactly once and a matched control utterance with no
internal repetition, each concatenated $k\in\{2,4,8,16,32\}$ times with $0.25\,$s silences: the true count is exact
by construction, and both recognisers score identical audio. The replacement, a wav2vec2 recogniser (large, LV-60
self-trained) under connectionist temporal classification (CTC), has no autoregressive decoder and no internal
language model: per-frame character distributions, purely local, so errors cannot correlate with repetition.
\begin{table*}[tp]
\centering
\caption{Head-to-head validation on identical synthetic ground truth. Median counted/true ratio, $k\ge4$
pooled. The repeated-sentence arm is periodic at the sentence grain, not an aperiodic control.}
\label{tab:ctcval}
\small
\begin{tabular}{@{}llrl@{}}
\toprule
Recogniser & Material & $k\ge4$ ratio & Notes \\
\midrule
CTC & periodic (repeated word) & 1.00 & 22/24 exact; 2 off by 1 ($k{=}16,32$) \\
CTC & repeated sentence (control) & 1.00 & 23/24 exact; 1 off by 1 ($k{=}32$) \\
Whisper large-v3 & periodic (repeated word) & 0.19 & no transcript at $k{=}16,32$ \\
Whisper large-v3 & repeated sentence (control) & 0.25 & no transcript at $k{=}16,32$ \\
\bottomrule
\end{tabular}
\end{table*}

\noindent Whisper's $k\ge4$ ratios pool $k{=}4$ and $k{=}8$ only: every $k{=}16$ and $k{=}32$ call returned no
transcript, a missing measurement, not a low ratio.

\paragraph{The audit enlarged from one donor system to six.}
Table~\ref{tab:ctcval} is the original audit, six donor utterances from one source model. The enlarged audit takes
donor atoms from all six checkpoints, gated on presence (the CTC recogniser finds the target exactly once in the
unconcatenated clip): $27$ of $36$ atoms pass, giving $108$ trials at $k\ge4$. There the CTC judge's median ratio is
$1.00$ on every one of the six donors; Whisper scores $0.25$ on the $48$ trials that return a transcript and nothing
on the other $60$ of $108$. The main text quotes this audit, so the $0.19$ above and its $0.25$ are two audits, not
two answers.

\subsection{Four independent judges on the full population}\label{sec:judgefloor}
The remaining objection is that CTC's blank/repeat collapse is exactly the confound that would produce the headline
dissociation were the judge doing the counting. The same generated audio was therefore re-scored, under the paper's
own scoring and population rules, by four judges: the primary (wav2vec2 large, LV-60 self-trained), HuBERT large
LS-960 (same output structure, different pretraining), wav2vec2 robust-ft, and Whisper large-v3, autoregressive with
no blank-collapse rule at all.
\begin{table*}[tp]
\centering
\caption{Four judges on the full panel population: exact rates and repeated-versus-control gap in points
($k\ge6$, paired panel), then the gap by checkpoint and by family.}
\label{tab:fourjudges}
\small
\begin{tabular}{lrrrr}
\toprule
 & primary & HuBERT & robust-ft & Whisper \\
\midrule
control exact (\%) & 94.3 & 89.5 & 82.1 & 92.9 \\
repeated exact (\%) & 18.2 & 17.9 & 17.5 & 17.1 \\
gap (points) & $+76.7$ & $+72.3$ & $+65.1$ & $+75.5$ \\
\midrule
Llasa-1B & 78.1 & 78.6 & 65.2 & 69.9 \\
Llasa-3B & 77.2 & 74.8 & 65.9 & 76.0 \\
Llasa-8B & 81.4 & 80.3 & 69.7 & 71.7 \\
Qwen3-TTS-0.6B & 92.2 & 87.8 & 82.2 & 92.2 \\
Qwen3-TTS-1.7B & 60.0 & 57.8 & 50.0 & 71.1 \\
XTTS-v2 & 71.1 & \textbf{54.4} & 57.8 & 72.2 \\
\midrule
family: Llasa & 78.9 & 77.9 & 66.9 & 72.5 \\
family: Qwen3-TTS & 76.1 & 72.8 & 66.1 & 81.7 \\
family: XTTS & 71.1 & \textbf{54.4} & 57.8 & 72.2 \\
\bottomrule
\end{tabular}
\end{table*}

\noindent Every judge's gap is positive in $6$ of $6$ checkpoints and $3$ of $3$
families, clearing the pre-committed $+50$-point floor
(Table~\ref{tab:fourjudges}); the weakest judge keeps $85\%$ of the
published gap. Decisive is which arm moves: across the four judges the
repeated arm's exact rate spans $1.0$ point and the control arm's $12.2$
(top rows of the table); changing the scorer moves the arm the models get
right and pins the arm they get wrong, the reverse of a judge-side
collapse. The disagreement asymmetry agrees ($+0.27$ counts where the
confound requires $-1.0$ or lower), and the one CTC-specific confound runs
in our disfavour: blank-collapse merges back-to-back identical words
(``VERYVERY VERRYYVERYVERY'' for twelve requested copies of \emph{very}),
depressing same-word runs and never distinct-word controls, so it can
inflate the gap but not manufacture it.

\paragraph{A cross-lingual check.}\label{sec:spanish}
The audit replicates in Spanish (same XTTS-v2 weights): on Spanish ground
truth built the same way, the CTC judge scores $1.00$ of the true count
where Whisper large-v3 scores $0.19$ and returns nothing at the two
highest $k$. The Spanish counting arm itself failed its pre-registered
rendering-quality gate (control exact rate $20.4\%$ against a $50\%$
floor), so the cross-lingual dissociation is reported as untested rather
than replicated.

\section{Checkpoint-level inference and the specification curve}
\label{sec:inference}

The exact-rate gap is positive in six of six checkpoints and never below $60$ points. The
checkpoint is the sampling unit, one repeated-minus-control difference per contrast, positive
meaning present; Table~\ref{tab:percheck} gives the per-checkpoint contrasts,
Table~\ref{tab:paneltests} the panel summary, and the exact-rate column feeds
the main text's headline table.

\begin{table*}[tp]\centering
\caption{The three per-checkpoint contrasts, signed so positive means the deficit is present: exact-rate and median-error gaps in percentage points, the capacity gap in distinguishable states per unit $\log k$.}
\label{tab:percheck}
\begin{tabular}{lrrr}
\toprule
checkpoint & exact-rate (pts) & median-error (pts) & capacity \\
\midrule
Llasa-1B~\cite{ye2025llasa} & +78.1 & +12.5 & +38.5 \\
Llasa-3B & +77.2 & +8.3 & +46.5 \\
Llasa-8B & +81.4 & +16.7 & +30.5 \\
Qwen3-TTS-0.6B~\cite{hu2026qwen3tts} & +92.2 & +12.5 & +14.0 \\
Qwen3-TTS-1.7B & +60.0 & +0.0 & +8.1 \\
XTTS-v2~\cite{casanova2024xtts} & +71.1 & +12.5 & +10.7 \\
\bottomrule
\end{tabular}
\end{table*}

\begin{table*}[tp]\centering
\caption{Panel-level summary of the three contrasts: mean, bootstrap 95\% CI, two-sided exact sign-test $p$, and the count of positive checkpoints.}
\label{tab:paneltests}
\begin{tabular}{lrrrl}
\toprule
contrast & mean & 95\% CI & $p$ & positive \\
\midrule
exact-rate gap (pts) & 76.7 & [68.4, 84.4] & 0.031 & 6/6 \\
median-error gap (pts) & 10.4 & [5.6, 13.9] & 0.062 & 5/6 \\
capacity gap & 24.7 & [13.7, 36.5] & 0.031 & 6/6 \\
\bottomrule
\end{tabular}
\end{table*}

\noindent\textbf{No multiplicity correction can rescue these $p$-values; we do not ask it to.}
The smallest two-sided exact $p$ at $n=6$ is $2/2^6=0.031$ and Holm's floor $3\times0.031=0.094$,
so the paper rests on effect sizes and consistency; direction generalises, magnitude does not.

\subsection{The clustering, modelled instead of counted}

Modelled any way that respects the clustering, the arm effect stays near $76$ points and every
interval excludes zero. Six checkpoints are three families, so independent replicates would
double-count; gaps recomputed from the shared panel definition (Section~\ref{sec:shape}) match
the numbers above exactly.

\paragraph{Mixed-effects and cluster-robust models.} The primary specification, crossed random
intercepts with a family-varying arm slope, pays its standard error out of three families, not
$982$ generations; Table~\ref{tab:hier-intervals} gives every construction with its clustering
unit.

\begin{table*}[tp]
\centering
\caption{Interval constructions for the arm effect (control $-$ repeated,
percentage points). The Wald row is reported for reference and is \emph{not} one
of the six intervals the main text's ``all six'' refers to.}
\label{tab:hier-intervals}
\small
\begin{tabular}{@{}lrrrl@{}}
\toprule
Construction & $n$ clusters & Estimate & 95\% interval & Criterion \\
\midrule
Mixed model, crossed RE, Wald (reference only) & 3 fam. & 76.4 & [71.9, 80.9] & $z$ \\
\addlinespace
\multicolumn{5}{@{}l}{\emph{the six constructions the main text's ``all six'' counts:}} \\
Mixed model, crossed RE, $t(2)$          & 3 fam.  & 76.4 & [66.6, 86.3] & $t(2)$ \\
OLS, clustered by family                 & 3       & 76.2 & [68.7, 83.7] & $t(2)$ \\
OLS, clustered by checkpoint             & 6       & 76.2 & [63.2, 89.2] & $t(5)$ \\
Family-level means, $t$-interval         & 3       & 75.4 & [65.6, 85.2] & $t(2)$ \\
Cluster bootstrap over families (20k)    & 3       & 76.0 & [71.1, 79.0] & percentile \\
Cluster bootstrap over checkpoints (20k) & 6       & 76.2 & [67.2, 84.8] & percentile \\
\midrule
\multicolumn{5}{@{}l}{\emph{check, not one of the six (different link function):}} \\
GEE, binomial, clustered by family & 3 & \multicolumn{2}{c}{OR $\approx$ 96, log-odds $[3.31,5.83]$} & $t(2)$ \\
\bottomrule
\end{tabular}
\end{table*}

The main text quotes the widest of the six, the checkpoint-clustered OLS; all six exclude zero,
and the GEE agrees on its different scale. \textbf{The Wald interval is excluded} as artefact:
three families barely identify the family-varying arm-slope variance, so shrinkage undercuts
the between-family spread.

\paragraph{A hierarchical Bayesian model.} A binomial logit model, partially pooled: control
and gap logits each carry a grand mean, family offset and nested checkpoint offset. Two
independent samplers converge and agree to $0.12$ points on the panel gap
(Table~\ref{tab:hier-bayes}).

\begin{table}[htbp]
\centering
\caption{Posterior arm-effect gap (percentage points, control $-$ repeated),
partially pooled; all $P(\mathrm{gap}>0)=1.000$ to the precision shown.
Llasa~\cite{ye2025llasa}, Qwen3-TTS~\cite{hu2026qwen3tts},
XTTS-v2~\cite{casanova2024xtts}.}
\label{tab:hier-bayes}
\small
\begin{tabular}{@{}llrr@{}}
\toprule
Level & Unit & Mean & 95\% credible interval \\
\midrule
Family & Llasa      & 79.3 & [73.4, 84.4] \\
Family & Qwen3-TTS  & 74.1 & [67.7, 79.9] \\
Family & XTTS       & 73.7 & [63.5, 82.2] \\
\addlinespace
Checkpoint & Llasa-1B & 79.5 & [70.5, 86.9] \\
Checkpoint & Llasa-3B & 77.2 & [67.0, 85.5] \\
Checkpoint & Llasa-8B & 81.0 & [72.7, 88.0] \\
Checkpoint & Qwen3-TTS-0.6B & 87.6 & [80.1, 93.8] \\
Checkpoint & Qwen3-TTS-1.7B & 60.5 & [50.4, 70.0] \\
Checkpoint & XTTS-v2 & 73.7 & [63.5, 82.2] \\
\bottomrule
\end{tabular}
\end{table}

With random effects at zero the \emph{panel} gap is $74.9$ points, $[41.5, 89.4]$, beneath the
intervals above because it marginalises over the between-family variance they condition away.
The main text leads with the posterior \emph{predictive} gap for a new family
(Table~\ref{tab:hier-prior}), wide enough to place the effect with high probability, not to
exclude a deficit-free fourth architecture.

\paragraph{Prior sensitivity, both directions.} The widest prior's predictive interval includes
zero, a one-in-twenty-five chance that a fourth family shows no deficit or reverses it, the cost
of locating a between-family variance from three families; every prior leaves a large effect in
every observed family. Table~\ref{tab:hier-prior} refits the identical model under tighter and
much wider priors.

\begin{table*}[tp]
\centering
\caption{Prior sensitivity on the four group-level scales $\sigma$: new-family
predictive gap (points).}
\label{tab:hier-prior}
\small
\begin{tabular}{@{}lrrr@{}}
\toprule
Prior on $\sigma$ & Mean & 95\% CrI & $P(\mathrm{gap}>0)$ \\
\midrule
HalfNormal(0.5), tighter, favourable & 71.7 & [26.6, 90.9] & 1.000 \\
HalfNormal(1.0), reported & 65.3 & [5.9, 93.3] & 0.996 \\
HalfNormal(3.0), much wider, unfavourable & 52.3 & [$-$1.6, 96.1] & 0.956 \\
\bottomrule
\end{tabular}
\end{table*}

\paragraph{The specification curve.} $420$ specifications, zero sign reversals
(Table~\ref{tab:hier-spec}). Six crossed axes: exclusion rule, $k$ threshold, cycled versus
never-cycled (aperiodic) controls, seed subset, aggregation level, and outcome, exact-rate
versus median-relative-error gap, scales reported separately; of $600$ crossed cells, $180$
duplicate populations, leaving $420$, none infeasible.

\begin{table*}[tp]
\centering
\caption{Specification curve; $n$ is the number of specifications under that
outcome definition.}
\label{tab:hier-spec}
\small
\begin{tabular}{@{}lrrrrr@{}}
\toprule
Outcome & $n$ & Range (pts) & Median & Positive & Zero / negative \\
\midrule
Exact-rate gap & 210 & [34.4, 86.4] & 70.5 & 210/210 & 0 / 0 \\
Median-error gap & 210 & [0.0, 19.5] & 8.3 & 197/210 & 13 / 0 \\
\midrule
Both outcomes combined & 420 & --- & --- & 407/420 & 13 / 0 \\
\bottomrule
\end{tabular}
\end{table*}

The exact-rate spread is $2.5\times$ and never changes sign; the median-error gap is never
negative, its non-positive cells exactly zero. The main text's $0$ of $420$ is exactly
sign-preservation; the two ranges must not be averaged.

\begin{table*}[tp]
\centering
\caption{Which forking choice moves the exact-rate gap (median gap at each
level, percentage points), sorted by spread, descending.}
\label{tab:hier-axis}
\small
\begin{tabular}{@{}l r p{6.6cm}@{}}
\toprule
Axis & Spread (pts) & Median gap by level \\
\midrule
$k$ threshold & 22.8 & $k{\ge}1$: 58.0, $k{\ge}6$: 74.7, $k{\ge}8$: 78.3, $k{\ge}12$: 80.7, $k{\ge}16$: 69.4 \\
Exclusion rule & 9.7 & panel: 75.8, +templates: 68.8, +cap-hits: 76.7, +degenerate: 75.7, everything: 67.1 \\
Aggregation level & 2.3 & pooled: 70.6, checkpoint: 71.3, family: 69.1 \\
Control construction & 2.3 & cycled: 72.0, aperiodic: 69.7 \\
Seed subset & 1.1 & all seeds: 69.7, seed 0 only: 70.8 \\
\bottomrule
\end{tabular}
\end{table*}

The $k$ threshold owns the spread (Table~\ref{tab:hier-axis}): $k_{\min}=1$ pulls in the
sub-horizon range where the arms have not yet dissociated, compressing the gap mechanically. The
weakest cell keeps every degenerate and cap-hit generation and stays $34$ points positive.

\subsection{What the brackets mean}
\label{sec:brackets}

Every bracketed pair is a $95\%$ interval, the kind set by the estimand: Wilson score for rates,
nonparametric bootstrap over templates or items for fitted quantities, the across-checkpoint
range for the $n=6$ panel means. Item-pooled intervals ignore within-checkpoint clustering and
run too narrow; the main text leans on the checkpoint-level number.

\section{The shape of the deficit}
\label{sec:shape}

\paragraph{Saturating versus proportional.} Proportional wins in five of six checkpoints and
pooled, by orders of magnitude in residual: no counting horizon within $k\le32$. The refuted
contraction hypothesis (Section~\ref{sec:contraction}) would bound the largest separable
repetition count by a constant $N^\ast$, so a decoder driven past that horizon should
\emph{saturate} whatever the request; median rendered count above $k=6$ is fitted with a
constant $a$, a proportional $bk$, and the interpolating $N^\ast(1-e^{-k/N^\ast})$, free to put
a horizon anywhere (Table~\ref{tab:satfits}).

\begin{table*}[tp]\centering
\caption{Constant, proportional and soft-horizon fits to the median rendered count above $k=6$: residual sums of squares, fitted slope $b$, fitted horizon $N^\ast$, and the winning form.}
\label{tab:satfits}
\begin{tabular}{lrrrrl}
\toprule
model & sat.\ SSE & prop.\ SSE & $b$ & soft $N^\ast$ & best \\
\midrule
Llasa-1B~\cite{ye2025llasa} & 477.5 & 35.0 & 0.855 & 85 & proportional \\
Llasa-3B & 464.2 & 79.3 & 0.913 & 119 & soft horizon \\
Llasa-8B & 353.5 & 2.0 & 0.830 & 71 & proportional \\
Qwen3-TTS-0.6B~\cite{hu2026qwen3tts} & 499.3 & 1.4 & 0.953 & 338 & proportional \\
Qwen3-TTS-1.7B & 668.8 & 5.1 & 1.107 & 640 & proportional \\
XTTS-v2~\cite{casanova2024xtts} & 378.0 & 0.8 & 0.860 & 90 & proportional \\
\midrule
pooled & 485.3 & 2.6 & 0.950 & 278 & proportional \\
\bottomrule
\end{tabular}
\end{table*}

\noindent The soft-horizon fit, free to find a plateau, puts $N^\ast$ at or past the ladder's
top in all six. That does not bear on the refuted contraction account either way: the
ladder stops short of the regime a horizon would govern.

\paragraph{Which rows the panel may report.} Four exclusions, one shared population definition.

\begin{enumerate}
\item \textbf{Ablation arms are not panel members:} the XTTS-v2 decoding-ablation arms
(Section~\ref{sec:penalty}) are one checkpoint under altered decoding; pooling would count XTTS-v2
five times.
\item \textbf{Degenerate and empty audio has no count}: reported as its own rate, not as a large
negative error.
\item \textbf{Templates whose vocabulary the judge cannot transcribe:} the CTC judge
(Section~\ref{sec:judge}) renders neither t2 filler, so that template scored the recogniser's
orthography; hitting both arms, the criterion cannot favour the control side.
\item \textbf{Items cut off by our own token budget:} generations that hit the harness limit have
median relative error $-0.44$ against $-0.08$, our budget truncating audio, not failed counting;
removal leaves the pooled median unchanged, per-model figures not.
\end{enumerate}

A fifth correction is a scoring fix, not an exclusion: the unit counter stopped at the first unit
it could not find, right for repeated items, wrong for controls; it now skips, verified
bit-identical on repeated items, lifting pooled control mean error $-0.150\to-0.090$.

\paragraph{The extension ladder.} Because the main ladder cannot tell a distant horizon from
none, the same three carriers ran at $k\in\{48,64,96,128\}$, with two recorded protocol
departures: the token budget rises $2048\to8192$, and the controls draw from a $146$-word pool
never cycled. XTTS-v2 is not run: every $k\ge48$ item exceeds its decoder's ceiling, censoring
counts by architecture, not horizon.

\paragraph{Do the exclusions fall evenly on the two arms?} The one unbalanced rule removes the
worst repeated generations, shrinking the reported gap. Table~\ref{tab:exclarms} gives panel rates
before any other filter.

\begin{table*}[tp]\centering
\caption{Exclusion rates by arm, before any other filter.}
\label{tab:exclarms}
\begin{tabular}{lrrr}
\toprule
arm & template rule & token budget & degenerate \\
\midrule
repeated ($n=1080$) & 16.7\% & 10.5\% & 4.4\% \\
control ($n=972$)   & 16.7\% & 5.6\%  & 3.7\% \\
\bottomrule
\end{tabular}
\end{table*}

The template rule is exactly balanced, structurally. The budget rule hits repeated items about
twice as often, as a looping model should on periodic text; excluding them shrinks the gap,
keeping them would enlarge it. Degeneracy is near-balanced and small on both arms.

\subsection{Fitted-form dependence of the horizon ratio}
\label{sec:horizonform}

The horizon ratio's direction is form-independent: no curve family moves any checkpoint across
$1$. The saturating form was adopted only after the main ladder ran proportional, so the refit
uses families we did not choose: one-parameter, unit slope at $k\to0$, asymptoting to $\hat K$,
differing only in the bend.

\noindent\textbf{A flagged repair, favourable to us:} the Qwen harness counted decode steps one
token short, leaving budget-truncated items inside the horizon fits; repaired and rerun, the
soft-horizon ratio moves $3.5\to4.2$ and the three-family range $2.7$--$4.3\to3.1$--$5.4$, only
the two Qwen checkpoints moving (Table~\ref{tab:formfams}).

\par\medskip\noindent\begin{minipage}{\columnwidth}
\inlinecaption{Three saturating families refit to the pooled extension ladder: fitted horizons for each arm and their ratio.}\label{tab:formfams}
\centering
\begin{tabular}{llrrr}
\toprule
family & $\hat c(k)$ & $\hat K_{\mathrm{rep}}$ & $\hat K_{\mathrm{ctl}}$ & ratio \\
\midrule
soft horizon & $\hat K(1-e^{-k/\hat K})$ & 24.8 & 104.8 & 4.22 \\
hyperbolic   & $\hat K k/(\hat K+k)$     & 33.6 & 181.7 & 5.40 \\
tanh         & $\hat K\tanh(k/\hat K)$   & 23.7 &  74.6 & 3.14 \\
\bottomrule
\end{tabular}
\end{minipage}\par\medskip

\noindent The soft horizon is the reported fit, hyperbolic and tanh bracketing it;
$\hat K_{\mathrm{ctl}}$ is unchanged to the digit under the repair, matching a bug confined to
the repeated arm's truncated items. Table~\ref{tab:formper} repeats it per
checkpoint.

\par\medskip\noindent\begin{minipage}{\columnwidth}
\inlinecaption{Per-checkpoint horizon ratios under the three saturating families.}\label{tab:formper}
\centering
\begin{tabular}{lrrr}
\toprule
checkpoint & soft horizon & hyperbolic & tanh \\
\midrule
Llasa-1B~\cite{ye2025llasa} &  5.97 &  8.46 & 4.50 \\
Llasa-8B &  4.12 &  5.23 & 3.06 \\
Qwen3-TTS-0.6B~\cite{hu2026qwen3tts} & 13.13 & 16.33 & 4.90 \\
Qwen3-TTS-1.7B &  0.29 &  0.57 & 0.07 \\
\bottomrule
\end{tabular}
\end{minipage}\par\medskip

\noindent\textbf{Direction is form-independent; magnitude is not.} All three families agree on
which checkpoints lower the horizon, the one below $1$ being Section~\ref{sec:weakest}'s
exception; the headline $4.2$ is a point inside $3.1$--$5.4$, so the main text quotes the range
beside it. This tests the curve choice only: every family saturates by construction.

\subsection{Do our own exclusions make the gap?}
\label{sec:exclusions}

Our exclusions flatter the gap by about fourteen points and do not create it; the main text
gives both numbers. About a quarter of panel generations are removed by three rules, each
defensible alone: three defensible rules can still sum to a selected sample.
Table~\ref{tab:exclsens} restores each rule in turn.

\begin{table*}[tp]\centering
\caption{The gap with each exclusion rule lifted in turn, and with nothing excluded at all.}
\label{tab:exclsens}
\begin{tabular}{lrrrr}
\toprule
population & $n$ & exact rep & exact ctl & gap \\
\midrule
panel (as reported)   & 1234 & 18.2\% & 94.3\% & 76.1 \\
$+$ judge-unmeasurable template & 1424 & 17.2\% & 79.8\% & 62.6 \\
$+$ budget-truncated items & 1362 & 16.3\% & 93.1\% & 76.7 \\
$+$ degenerate audio (scored 0) & 1242 & 17.9\% & 93.9\% & 76.0 \\
nothing excluded at all & 1620 & 14.8\% & 76.5\% & 61.7 \\
\bottomrule
\end{tabular}
\end{table*}

\noindent Only the template rule does work, and on the \emph{control} side: the judge emits
neither t2 filler, so restored t2 rows score the recogniser's vocabulary, judge noise on the
control arm shrinking the gap. The budget rule, the one that could have flattered us, does not,
truncation hitting both arms; degenerate audio scored zero is the harshest choice.

\subsection{The inverting checkpoint is the panel's weakest}
\label{sec:weakest}

Qwen3-TTS-1.7B~\cite{hu2026qwen3tts}, the one checkpoint that shows no repeated-side saturation
within $k\le128$ and inverts the horizon ratio, is also last on all three measures in
Table~\ref{tab:percheck}, and the only checkpoint last on any: the exception sits
where the effect is smallest, where smallest tips into absent. Separating its two candidates, a
per-repetition map that may not contract (Section~\ref{sec:contraction}) or a horizon past
$k=128$, needs a longer ladder than its generation budget allows. A weakest checkpoint exists
in any panel, so the ranking is no evidence for the contraction account; but it rules out the worry that
the horizon result rests on discarding a contradicting checkpoint.

\section{Alternative explanations, answered with measurements}
\label{sec:alternatives}

Five falsifiable alternative explanations, each answered with a measurement, none overturning
the dissociation; the closing paragraph collects the statistical machinery, including an
uncorrected count of the comparisons.

\paragraph{Naturalness.} The control is the \emph{less} probable text under every independent
scorer: naturalness runs opposite the alternative. \textbf{Objection:} twelve copies of
\emph{very} are far more out-of-distribution than twelve distinct intensifiers
(Section~\ref{sec:testset}), so failure driven by improbability rather than periodicity would
leave the repeated item less probable in most of the 54 pairs. \textbf{Test:} per-token
cross-entropy, same 54 pairs, three scorers in the roles Table~\ref{tab:naturalness} names; the
template-bootstrap gap intervals exclude zero for both independent scorers.

\begin{table*}[tp]
\centering
\caption{Naturalness check, three scorers, 54 matched pairs. ``Ctl.\ higher'': fraction of pairs in
which the control has the higher (less probable) per-token cross-entropy.}
\label{tab:naturalness}
\small
\begin{tabular}{@{}lp{2.6cm}rrrr@{}}
\toprule
Scorer & Role & Rep.\ mean & Ctl.\ mean & Gap & Ctl.\ higher \\
\midrule
Llasa-1B & panel backbone & 14.85 & 16.05 & 1.20 & 89\% \\
phi-2 & independent, primary & 3.69 & 4.89 & 1.20 & 87\% \\
gpt2-large & secondary cross-check & 3.36 & 5.34 & 1.98 & 100\% \\
\bottomrule
\end{tabular}
\end{table*}

\noindent\textbf{The disagreement.} Scorers agree on direction and pooled size, not on shape
across $k$: phi-2's gap narrows and reverses at $k\ge24$, wholly past every panel $k^\ast$ of
1--8 (Table~\ref{tab:unitinv}). Where collapse happens both agree, and only that conjunction is
claimed; the harder-to-predict member, always the control, is rendered correctly.

\paragraph{Unit invariance.} Collapse points differ by a mean 1.7 repetitions across units while
token cost at collapse differs 1.34$\times$: repetition's pattern, not the budget's.
\textbf{Objection:} a purely acoustic attractor predicts sentence repetition, several times
costlier in tokens per repetition, collapsing at much smaller $k$; the contraction account
predicts comparable $k^\ast$ across units. \textbf{Test} (Table~\ref{tab:unitinv}): per model,
$k^\ast$, the largest $k$ with accuracy above 0.5, for both units.

\begin{table*}[tp]
\centering
\caption{Unit invariance, as committed. Token counts are medians at the reported $k^\ast$; n/a where no
$k$ cleared the $0.5$ threshold.}
\label{tab:unitinv}
\small
\begin{tabular}{@{}lrrrr@{}}
\toprule
Model & $k^\ast_{\mathrm{word}}$ & tok.\ (word) & $k^\ast_{\mathrm{sent}}$ & tok.\ (sent) \\
\midrule
Llasa-1B & 0 & n/a & 3 & 351 \\
Llasa-3B & 1 & 209 & 3 & 457 \\
Qwen3-TTS-0.6B & 4 & 53 & 4 & 80 \\
Qwen3-TTS-1.7B & 8 & 80 & 4 & 86 \\
XTTS-v2 & 4 & 89 & 4 & 128 \\
XTTS-v2 (no penalty, ablation) & 2 & 63 & 1 & 29 \\
\bottomrule
\end{tabular}
\end{table*}

\noindent Llasa-8B, completed after the table was committed, returns
$k^\ast_{\mathrm{word}}{=}4$, $k^\ast_{\mathrm{sent}}{=}3$, leaving the
mean difference $1.7$ unchanged; the token ratio, $1.50\times$, stays
provisional. These $k^\ast$ are Whisper-judged
(Section~\ref{sec:judge}); the conclusion needs only a consistent
threshold across families.

\paragraph{The capacity confound.} Raw, diversity-adjusted, and correct-only panel medians agree
within two hundredths (Table~\ref{tab:capconf}): the rank gap is not monotonous audio.
\textbf{Objection:} sixteen correct \emph{very}s make audio genuinely more monotonous than
sixteen distinct words; effective-rank decline could measure that surface fact. It holds only
if the gap in rank growth closes once output diversity is regressed out and only correct
renderings are compared. \textbf{Test:}
$\mathcal{N}_{\mathrm{eff}}$ regressed on $\log k$ with three diversity covariates of the
emitted token sequence, held only for Llasa, so the other adjusted cells are inestimable, not
zero; and refitted on items labelled correct (Section~\ref{sec:scoring}).

\begin{table*}[tp]
\centering
\caption{Capacity-confound ratios (repeated/control slope of $\mathcal{N}_{\mathrm{eff}}$ against
$\log k$). $n$ pairs are repeated-side/control-side rows entering each regression; --- marks an
inestimable cell (fewer than 12 usable rows, or fewer than 4 distinct $k$ values).}
\label{tab:capconf}
\small
\begin{tabular}{@{}lrrr@{}}
\toprule
Model & Raw & +diversity & correct-only \\
\midrule
Llasa-1B & 0.42 & 0.50 & 0.01 ($n{=}14$/32) \\
Llasa-3B & 0.28 & 0.34 & 0.80 ($n{=}24$/22) \\
Llasa-8B & 0.43 & 0.75 & --- ($n{=}0$/0) \\
Qwen3-TTS-0.6B & 0.57 & --- & 0.31 ($n{=}42$/84) \\
Qwen3-TTS-1.7B & 0.78 & --- & 0.47 ($n{=}64$/82) \\
XTTS-v2 & 0.60 & --- & 0.56 ($n{=}22$/41) \\
\midrule
\textbf{Panel median} & \textbf{0.50} & \textbf{0.50} & \textbf{0.47} \\
XTTS-v2 (no penalty, abl., not pooled) & 0.89 & --- & --- ($n{=}9$/22) \\
\bottomrule
\end{tabular}
\end{table*}

One flag: Llasa-1B's correct-only 0.01 is one noisy near-zero slope driving a ratio to zero
without the gap closing. Two qualifications stand: the adjusted control is well-powered for
only three of six checkpoints, and the correct-only control corroborates directionally, not
decisively; its Whisper-judged label (Section~\ref{sec:judge}) only removes items from the
repeated correct pool, leaving cells under-powered, not inflated.

\paragraph{Count survival in the representation.} Repeated retention falls below its control in
six of six checkpoints (Table~\ref{tab:proberet}). \textbf{Objection:} in text-only autoregressive
LMs a linear probe, exactly the Lipschitz readout the contraction account constrains, decodes
the true count near-perfectly at every layer: unused, not
erased~\cite{venkatesh2026repeated}; degradation predicts the opposite. \textbf{Test:} ridge
regression predicts $\log_2 k$ from the mean hidden state over the first and last thirds of
each instrumented trajectory, with leave-one-template-out cross-validation; retention,
best-layer late-window over early-window $R^2$, cancels probe capacity and item difficulty,
the matched control decoding the same quantity from non-repetitive text of equal length.
Best-layer $R^2$ stays modest, far from the text-LM near-perfect case.

\begin{table*}[tp]
\centering
\caption{Ridge-probe retention (late-window best-layer $R^2$ / early-window best-layer $R^2$). All
entries $n{=}54$ items.}
\label{tab:proberet}
\small
\begin{tabular}{@{}lrr@{}}
\toprule
Model & Repeated retention & Control retention \\
\midrule
Llasa-1B & 0.97 & 1.09 \\
Llasa-3B & 0.87 & 0.99 \\
Llasa-8B & 0.96 & 1.09 \\
Qwen3-TTS-0.6B & 1.46 & 1.74 \\
Qwen3-TTS-1.7B & 0.77 & 1.12 \\
XTTS-v2 & 1.92 & 2.85 \\
\midrule
\textbf{Panel median} & \textbf{0.97} & \textbf{1.11} \\
XTTS-v2 (no penalty, abl., not pooled) & 1.37 & 1.40 \\
\bottomrule
\end{tabular}
\end{table*}

\noindent\textbf{A second pooling defect, found in verification.} The first-reported statistics
pooled the no-penalty ablation; the six panel checkpoints alone give the medians shown, repeated
still lower in six of six, the ablation a consistent, separate seventh point.
\textbf{Conclusion:} qualified favour for representational degradation over output policy; yet
below-1 retention is degradation, not erasure, and the accounts are not exclusive.

\paragraph{The XTTS repetition-penalty ablation.} \textbf{Objection and test:} XTTS-v2 ships a
5.0 repetition penalty on acoustic-token decoding (Section~\ref{sec:inference}); its high
$k^\ast$ could be that crutch's artifact. The ablation is the identical checkpoint at penalty
1.0, excluded from every panel-pooled statistic. \textbf{Result:} penalty off gives strictly
worse counting on both arms, a wider relative dissociation, and a compressed capacity range:
the paired capacity gap moves from excluding zero to crossing it, the one cell in the
panel-plus-ablation set where the capacity contrast misses significance, reported, not omitted;
by CTC relative count error both arms degrade too far to isolate periodicity from decoding
breakdown. The selective-masking reading is withdrawn; what stands: an ablation uninformative
about the dissociation, XTTS-v2 penalty-on still the conservative panel member.

\paragraph{Statistical treatment.} \textbf{Wilson intervals} for every proportion, working at
the 0\% and 100\% boundaries. \textbf{Bootstrap over templates, not items:} the 54 pairs are 6
templates $\times$ 9 values of $k$, so item resampling understates uncertainty; $n$ accompanies
every estimate. \textbf{No multiple-comparison correction; the counts, for readers to apply
their own:} six per-model capacity-gain tests, up to eighteen confound cells, six
probe-retention tests, nine naturalness $k$ values under two scorers; a family-wise guarantee
divides $\alpha{=}0.05$ by six for the primary contrast, or by the low twenties.

\section{Decoding-rule variations}
\label{sec:decoding}
\subsection{The repetition penalty}
\label{sec:penalty}

The strongest decoding-level rival, and the emptiest: three checkpoints never had a repetition
penalty and show a 78.9-point gap. The families ship penalties an order of magnitude apart
(XTTS-v2 5.0, Qwen3-TTS 1.05, Llasa none), and a penalty by construction suppresses repeated
tokens; had it produced the periodicity-specific deficit, the contraction account would explain what the
sampler already explains. Table~\ref{tab:penaltysweep} sweeps it an order of
magnitude in both directions.

\begin{table*}[tp]\centering
\caption{The repetition-penalty sweep on the two architectures that ship one; a collapsed control arm says nothing about repetition.}
\label{tab:penaltysweep}
\begin{tabular}{@{}lrrrl@{}}
\toprule
Architecture & Penalty & Rep.\ exact (\%) & Ctl.\ exact (\%) & Control \\
\midrule
XTTS-v2 & 1.00 & 15.3 & 30.9 & collapsed \\
XTTS-v2 & 2.00 & 15.6 & 92.2 & intact \\
XTTS-v2 & 3.00 & 21.1 & 91.1 & intact \\
XTTS-v2 (shipped) & 5.00 & 16.7 & 87.8 & intact \\
XTTS-v2 & 8.00 & 15.6 & 84.4 & intact \\
Qwen3-TTS-0.6B & 1.00 & 10.0 & 100.0 & intact \\
Qwen3-TTS-0.6B (shipped) & 1.05 & 7.8 & 100.0 & intact \\
Qwen3-TTS-0.6B & 1.50 & 10.0 & 80.0 & intact \\
Qwen3-TTS-0.6B & 3.00 & 16.7 & 83.3 & intact \\
\bottomrule
\end{tabular}
\end{table*}

\noindent\textbf{Sweeping it changes nothing that matters.} The two architectures disagree on
the slope's sign, the signature of a lever not acting on the relevant quantity; penalty
disabled, Qwen3-TTS-0.6B keeps a 90-point gap with no repetition penalty anywhere in the
decoder. Disabling XTTS-v2's penalty collapses its \emph{control}, so that arm says nothing
about repetition; slopes are fitted only over arms whose control survives, and Qwen3-TTS is the
useful contrast.

\subsection{Greedy decoding}
\label{sec:greedy}

Greedy decoding leaves the gap within three points of sampled: the dissociation lives in the
conditioning, not the draw. The contraction account bounds a readout with no reference to the
sampler, yet every panel generation was sampled; stochastic token choice is exactly the
perturbation breaking the posited time-invariant map, and greedy removes it. Table~\ref{tab:greedy}
compares the arms.

\par\medskip\noindent\begin{minipage}{\columnwidth}
\inlinecaption{Greedy against sampled decoding on Qwen3-TTS-0.6B, $k\ge6$.}\label{tab:greedy}
\centering
\begin{tabular}{@{}lrrrr@{}}
\toprule
Arm & $n$ & Exact rep.\ & Exact ctl.\ & Gap \\
\midrule
sampled, three seeds & 216 & 8.3\%  & 83.3\% & 75.0 \\
sampled, seed 0      &  72 & 5.6\%  & 83.3\% & 77.8 \\
greedy               &  72 & 11.1\% & 83.3\% & 72.2 \\
\bottomrule
\end{tabular}
\end{minipage}\par\medskip

\noindent Qwen3-TTS-0.6B, $k\ge6$; greedy is deterministic, one seed being the whole arm, and
the greedy-sampled pair is the one the main text's variations table quotes. \textbf{The
truncation confound, checked:} greedy on repetitive text hits a generation budget more readily,
and truncated counts are censored downward, exactly what would fake a greedy deficit; the
cap-hit rate is 0.0\% on both arms. \textbf{Limits:} one checkpoint, one alternative decoding
mode: Qwen3-TTS-0.6B's deficit is not a sampling artifact; the panel's could still be, and beam
search remains untested.

\subsection{Repetition-aware sampling}
\label{sec:ras}

The one mitigation purpose-built for this failure leaves the deficit intact, landing above the
pre-committed \textsc{survives} band. Sections~\ref{sec:greedy} and~\ref{sec:penalty} suppress a
token by its frequency in history with a scalar; VALL-E~2's Repetition Aware Sampling instead
``refines the original nucleus sampling process by accounting for token repetition in the
decoding history''~\cite{chen2024valle2}.

\noindent\textbf{What was implemented.} A reimplementation from the published description, no
code being released: draw by nucleus sampling; if the drawn token's repetition ratio over a
$K{=}10$-step window exceeds $t_r{=}0.1$, redraw from the untruncated distribution. Three arms
on Qwen3-TTS-0.6B: no-op threshold, paper defaults, and top-$p{=}0$. \textbf{Three gates:} at
the no-op threshold RAS reproduces stock decoding bitwise, the same decoder, not a similar one;
it fires on 48.8\% of repeated-item decode steps against 10.8\% on controls; and the control
arm holds 100.0\% exact, so the gap does not survive by the mitigation breaking the controls.

\noindent\textbf{Result: the deficit survives.} Exact-rate gaps, $k\ge6$, CTC judge, panel
exclusions (Table~\ref{tab:ras}).

\par\medskip\noindent\begin{minipage}{\columnwidth}
\inlinecaption{Repetition-aware sampling on Qwen3-TTS-0.6B: exact-rate gaps against the pre-committed band.}\label{tab:ras}
\centering
\begin{tabular}{@{}lrrl@{}}
\toprule
Arm & Rep.\ exact & Gap & 95\% CI \\
\midrule
no-op (= stock)          & 7.9\%  & $+92.1$ & $[+86.5, +97.8]$ \\
RAS, paper defaults      & 5.7\%  & $+94.3$ & $[+89.7, +98.9]$ \\
RAS, $\text{top-}p{=}0$  & 12.4\% & $+87.6$ & $[+80.9, +94.4]$ \\
\bottomrule
\end{tabular}
\end{minipage}\par\medskip

\noindent The band was pre-committed from this checkpoint's sweeps: \textsc{survives} at
$\ge +66.7$ points, the floor across penalty and greedy arms, \textsc{mitigates} at
$\le +33.4$; RAS lands above the band's top. \textbf{A reproduction for free:} the no-op arm
reproduces the published Qwen3-TTS-0.6B rows exactly, identical per-item counts on 233 of 233
paired items. \textbf{What weakens it:} one checkpoint; two top-$p$ settings; a description
implemented, not the authors' code; and a summation ambiguity in the published ratio, resolved
in the mitigation's favour, moving the fire rate by half but not the verdict.

\section{The period ladder in full}
\label{sec:period}

Exact renderings rise monotonically with period in every checkpoint,
and half the deficit survives at period $2$, where no token is
adjacent to itself. The published design never varied periodicity
alone, its repeated arm confounding period $1$ with verbatim token
identity; this ladder varies only the period, $p\in\{1,2,4,8,k\}$,
all thresholds pre-committed before any audio existed.

\paragraph{Design.} At fixed carrier, word count and requested count
$k\in\{16,24,32\}$: $p{=}1$ is $A\,A\,A\,A\,\dots$ (the published
repeated arm), $p{=}2$ is $A\,B\,A\,B\,\dots$, $p{=}4$ and $p{=}8$
cycle four and eight distinct words ($p{=}8$ the published cycled
control), and $p{=}k$ is $k$ distinct words (the published
never-cycled control). $p\in\{2,4,8\}$ draw rotation-balanced from
one shared eight-word pool and are the only vocabulary-matched arms;
$p{=}1$ and $p{=}k$ inherit the published arms' vocabularies, a
reported confound. Five carriers (the sixth is excluded panel-wide,
Section~\ref{sec:shape}), three seeds, three checkpoints (Llasa-1B,
Qwen3-TTS-0.6B, XTTS-v2), standard exclusions: $n=1134$ of $1395$
rows.

\paragraph{The full ladder.} Exact rate ($=k$ requested units delivered, in
order, the main text's statistic), headline ordered-scan rule, in
Table~\ref{tab:fullladder}.

\begin{table*}[tp]\centering
\caption{The full period ladder: exactly-right rate (\%) by period at $k\in\{16,24,32\}$ under the ordered-scan rule, with cell sizes.}
\label{tab:fullladder}
\begin{tabular}{@{}lrrrrrl@{}}
\toprule
checkpoint & $p{=}1$ & $p{=}2$ & $p{=}4$ & $p{=}8$ & $p{=}k$ & $n$ ($1,2,4,8,k$) \\
\midrule
Llasa-1B         &  3.6 & 54.6 & 83.3 &  92.5 & 75.0 & 28, 130, 84, 40, 44 \\
Qwen3-TTS-0.6B   &  6.8 & 41.3 & 73.3 & 100.0 & 86.7 & 44, 179, 90, 45, 45 \\
XTTS-v2          & 11.1 & 51.7 & 75.6 &  82.2 & 57.8 & 45, 180, 90, 45, 45 \\
\midrule
MEAN             &  7.2 & 49.2 & 77.4 &  91.6 & 73.2 & \\
\bottomrule
\end{tabular}
\end{table*}

\noindent Every checkpoint is monotone across $p=1,2,4,8$: a graded
deficit, not a step. $p{=}k$ sits below $p{=}8$ everywhere, as its
vocabulary caveat predicts, outside the monotonicity claim.

\paragraph{Five scoring rules.} $D(p)=E(8)-E(p)$ under five rules,
pre-committed calls $D(2)/D(1)\ge0.50$ PERIODICITY, $\le0.25$ VERBATIM
TOKEN IDENTITY, between GRADED; ratio pooled over checkpoints
(Table~\ref{tab:fiverules}).

\begin{table*}[tp]\centering
\caption{The period-2 recovery under five scoring rules: deficits $D(p)=E(8)-E(p)$ in points, the decisive ratio $D(2)/D(1)$, and the pre-committed verdict.}
\label{tab:fiverules}
\begin{tabular}{@{}lrrrrl@{}}
\toprule
rule & $D(1)$ & $D(2)$ & $D(4)$ & $D(2)/D(1)$ & verdict \\
\midrule
ordered scan (headline)     & 84.4 & 42.4 & 14.2 & 0.502 & PERIODICITY \\
unbounded recount           & 82.9 & 60.9 & 20.1 & 0.734 & PERIODICITY \\
strict conjunction          & 82.9 & 61.2 & 21.2 & 0.738 & PERIODICITY \\
cap hits kept (censored bound) & 81.9 & 43.5 & 13.5 & 0.531 & PERIODICITY \\
by family                   & 84.4 & 42.4 & 14.2 & 0.502 & PERIODICITY \\
\bottomrule
\end{tabular}
\end{table*}

\noindent The magnitude of $D(2)$ is rule-dependent, the PERIODICITY
verdict unanimous; ``by family'' reproduces the ordered scan, so four
independent methods, not five. The headline ratio clears $0.50$ by
only $0.002$, Qwen3-TTS-0.6B carrying the pooled call (per-checkpoint
scan ratios $0.426$, $0.630$, $0.430$), while both recount rules clear
$0.50$ on every checkpoint; the stored bootstrap interval was built on
strict conjunction, not the ordered scan, so the across-rule range is
the honest companion to the scan estimate.

\paragraph{$p{=}2$ by rotation.} Exact rate of the decisive cell over
the pool's four rotations (Table~\ref{tab:rotation}).

\par\medskip\noindent\begin{minipage}{\columnwidth}
\inlinecaption{The decisive $p{=}2$ cell over the pool's four rotations: exactly-right rate by rotation, and its range.}\label{tab:rotation}
\centering
\begin{tabular}{@{}lrrrrr@{}}
\toprule
checkpoint & rot.\ 0 & rot.\ 1 & rot.\ 2 & rot.\ 3 & range \\
\midrule
Llasa-1B         & 69.7 & 55.6 & 42.9 & 51.4 & 26.8 \\
Qwen3-TTS-0.6B   & 24.4 & 48.9 & 38.6 & 53.3 & 28.9 \\
XTTS-v2          & 57.8 & 55.6 & 53.3 & 40.0 & 17.8 \\
\bottomrule
\end{tabular}
\end{minipage}\par\medskip

\noindent A single-rotation $p{=}2$ arm could have read roughly $25\%$
to $70\%$ exact; rotation balancing is what makes the cell readable.

\paragraph{The disambiguation arm.} If the deficit were a failure to
individuate token-identical neighbours, marking each copy's boundary
should restore the count; punctuation restores $0.9\%$ of it, and
individuation is falsified. The arm holds period $1$ and edits only
the boundary: bare (\emph{very very very\ldots}, the published arm),
comma (\emph{very, very,\ldots}) and stop (\emph{Very. Very.\ldots}),
both length-matched, and and-conjoined (\emph{very and very\ldots}),
adding $k-1$ words; Qwen3-TTS-0.6B and XTTS-v2 only, $n=353$ after
exclusions. Pre-committed: recovered fraction
$R=(E(\text{dis})-E(\text{bare}))/(E(8)-E(\text{bare}))$ against each
checkpoint's own $E(8)$ ceiling, $R\ge0.50$ supported, $R\le0.15$
falsified. Exact rate by variant ($k\in\{16,24,32\}$; Qwen3 $=$
Qwen3-TTS-0.6B, XTTS $=$ XTTS-v2), with per-variant confound means
(Table~\ref{tab:disambig}).

\begin{table*}[tp]\centering
\caption{The disambiguation arm at period 1: exactly-right rate and recovered fraction $R$ by boundary variant, with per-variant confound means.}
\label{tab:disambig}
\begin{tabular}{@{}lrrrrrrr@{}}
\toprule
 & \multicolumn{2}{c}{exact rate} & \multicolumn{2}{c}{recovered $R$} & \multicolumn{3}{c}{confound means} \\
variant & Qwen3 & XTTS & Qwen3 & XTTS & words & chars & dur.\ (s) \\
\midrule
bare  &  6.8 & 11.1 & ---      & ---      & 31.8 & 170.6 & 11.65 \\
comma &  6.8 &  8.9 & $+0.000$ & $-0.031$ & 31.8 & 194.3 & 13.11 \\
stop  &  9.8 & 13.6 & $+0.031$ & $+0.036$ & 31.6 & 191.6 & 13.42 \\
and   & 15.6 & 11.1 & $+0.094$ & $+0.000$ & 54.8 & 263.0 & 16.23 \\
\bottomrule
\end{tabular}
\end{table*}

\noindent Mean $R$ over the length-matched cells is
$R_{\text{matched}}=0.009$, with $R_{\text{and}}=0.047$ beside it,
never averaged in: both far under the $0.15$ bar, the pre-committed
falsification. The and-variant nearly doubles the word count, which no
front-end can silently normalise away, so its small positive $R$ is
conservative, added length independently hurting counting.

\paragraph{The odd steps.} Periods $3$ and $6$, which neither a
power-of-two decoder nor a pool-cycling effect predicts should
interpolate, land exactly where periodicity predicts. Only $k{=}24$
admits such periods, so both ran there on Llasa-1B, Qwen3-TTS-0.6B,
Qwen3-TTS-1.7B and XTTS-v2, $720$ generations, all else as above;
pre-committed before any audio, against the $k{=}24$-only ladder:
\textsc{confirmed} iff $E(2)<E(3)<E(4)$ (Test~A) and $E(4)<E(6)<E(8)$
(Test~B) pooled, \textsc{refuted} iff either odd step reaches $E(8)$
or falls below $E(2)$.

Both tests pass: the pooled ordered-scan rates at $p=1,2,3,4,6,8$ are
the row the main text's period table carries, monotone across all six
periods under all four counting rules and every
leave-one-checkpoint-out fit. Every adjacent gap excludes zero except
the saturated ladder top (Table~\ref{tab:oddgaps}), so Test~B's upper
inequality is not separable from noise.

\begin{table}[htbp]
\centering
\caption{Adjacent-step gaps in exactly-right points on the $k{=}24$
ladder, carrier-cluster bootstrap 95\% intervals.}
\label{tab:oddgaps}
\small
\begin{tabular}{@{}lrr@{}}
\toprule
adjacent steps & gap (pts) & 95\% interval \\
\midrule
$E(3)-E(2)$ & $+10.9$ & $[+2.1,\,+16.8]$ \\
$E(4)-E(3)$ & $+12.9$ & $[+3.6,\,+21.2]$ \\
$E(6)-E(4)$ & $+14.7$ & $[+2.2,\,+28.3]$ \\
$E(8)-E(6)$ & $+2.7$  & $[-2.4,\,+7.7]$ \\
\bottomrule
\end{tabular}
\end{table}

\paragraph{Shuffled order at fixed composition.} Destroying the
periodic order while fixing each item's filler multiset recovers most
of the deficit on both Qwen checkpoints, none of it on Llasa-1B, and
destroys XTTS-v2: periodic ordering drives the deficit on some
architectures, not all. The test answers the rival the ladder cannot,
a decoder failing on low type-token ratio rather than periodic
ordering, per a pre-committed protocol. $p{=}2$ is uninformative by a
counting argument (exactly two orders of twelve A's and twelve B's avoid
adjacent equal tokens, both the periodic item, so aperiodicity forces
adjacent verbatim repetition) and at $p{=}8$ no deficit remains, so
everything rests on $p{=}4$, where zero adjacency is constructible and
was achieved exactly.

Pooled, shuffling recovers nothing: $E$ $75.2\% \to 75.3\%$ against a
$17.7$-point deficit, $R(4)=+0.006$, the pre-committed confirmation
band excluded ($P[\bar R \ge 0.50]=0.010$), written outcome
\textsc{intermediate}. But the pooled figure is arithmetic, not
consensus: three of four per-checkpoint intervals exclude zero, in
opposite directions (Table~\ref{tab:shuffled}).

\begin{table*}[tp]\centering
\caption{Shuffled order at fixed composition: exactly-right rate on periodic and shuffled $p{=}4$ items per checkpoint, with the bootstrap 95\% interval on the difference.}
\label{tab:shuffled}
\begin{tabular}{@{}lrrrl@{}}
\toprule
checkpoint & $E$ periodic & $E$ shuffled & $\Delta$ & 95\% CI \\
\midrule
Qwen3-TTS-1.7B & 80.0 & 97.8 & $+17.8$ & $[+5.6, +34.4]$ \\
Qwen3-TTS-0.6B & 73.3 & 91.1 & $+17.8$ & $[+6.7, +32.2]$ \\
Llasa-1B       & 74.1 & 68.9 & $-5.2$  & $[-22.0, +11.3]$ \\
XTTS-v2        & 73.3 & 43.3 & $-30.0$ & $[-53.3, -6.7]$ \\
\bottomrule
\end{tabular}
\end{table*}

\noindent Dropping XTTS-v2, $R(4)=+0.473$: a real architecture split,
reported as such, not as the mean. The collapse is no artifact: every
shuffled XTTS-v2 row renders clean, full-length speech, the collapse
survives the order-insensitive recount, and re-rendered periodic
controls are byte-identical to the published audio on all four
checkpoints, matching published rates on every cell. The placebo
correction, pre-committed to fire at $\Delta(8)\le-5$ points, tripped
by exactly $0.0$ and is declined; corrected or not, no conclusion
changes. The cost to the title: ``periodicity, not length'' holds here
on two checkpoints, fails on a third and reverses on a fourth, while
the length-matched dissociation holds on all six panel checkpoints and
a held-out architecture, so the title now carries what survives
everywhere (repetition, not length) and periodicity is the
architecture-dependent finding.

\section{The non-autoregressive baselines}
\label{sec:nonar}

One parallel decoder is immune to the deficit and one reproduces it,
so autoregression is not the cause: the architectural reading of the
deficit does not survive. The main text's scope claim is about
autoregressive decoders, and no non-autoregressive system had been
measured; two such baselines therefore run the published repeated and
control arms, beside the panel (Table~\ref{tab:nonar}).

\begin{table*}[tp]\centering
\caption{The published arms on the two non-autoregressive baselines, beside the panel: exactly-right rates and the gap in points.}
\label{tab:nonar}
\begin{tabular}{lrrr}
\toprule
model & repeated exact (\%) & control exact (\%) & gap (pts) \\
\midrule
Llasa-1B & 13.3 & 91.5 & +78.1 \\
Llasa-3B & 16.9 & 94.1 & +77.2 \\
Llasa-8B & 11.8 & 93.2 & +81.4 \\
Qwen3-TTS-0.6B & 7.8 & 100.0 & +92.2 \\
Qwen3-TTS-1.7B & 38.9 & 98.9 & +60.0 \\
XTTS-v2 & 16.7 & 87.8 & +71.1 \\
\midrule
\emph{F5-TTS (2024)} & 25.6 & 85.6 & +60.0 \\
\emph{VITS (2021)} & 65.6 & 58.9 & -6.7 \\
\bottomrule
\end{tabular}
\end{table*}

\noindent\textbf{The two disagree.} VITS shows no dissociation;
F5-TTS shows one close to the panel's, and neither has a
generation-step recurrence. Table~\ref{tab:kband} gives the gap per $k$ band,
in points.

\par\medskip\noindent\begin{minipage}{\columnwidth}
\inlinecaption{The gap per $k$ band, in points: the two non-autoregressive baselines against the autoregressive panel.}\label{tab:kband}
\centering
\begin{tabular}{lrrr}
\toprule
 & VITS & F5-TTS & AR panel \\
\midrule
$k=2$--$4$ & +6.7 & +8.9 & -1.0 \\
$k=6$--$8$ & -10.0 & +43.3 & +60.0 \\
$k=12$--$16$ & -13.3 & +56.7 & +85.7 \\
$k=24$--$32$ & +3.3 & +80.0 & +83.4 \\
\bottomrule
\end{tabular}
\end{minipage}\par\medskip

F5-TTS's gap grows with $k$ like the panel's, its durations scale with
$k$ (median $2.6$\,s at $k{=}1$, $11.6$\,s at $k{=}32$), so it renders
rather than truncates, and its controls hold above $83\%$:
periodicity-specific there too.

\noindent\textbf{VITS's null is not a floor effect.} At $k=6$--$8$
VITS renders repeated $76.7\%$ and control $66.7\%$ exact: far off the
floor, ordered the opposite way. Nor is it the stock configuration:
the stock duration predictor would have to carry the count (output
length is the sum of predicted phoneme durations), so it was perturbed
at two settings chosen to degrade it (Table~\ref{tab:vitscfg}).

\begin{table*}[tp]\centering
\caption{VITS under perturbed duration settings; the gap is control minus repeated, negative in every arm.}
\label{tab:vitscfg}
\begin{tabular}{llrrr}
\toprule
arm & setting & exact rep & exact ctl & gap \\
\midrule
stock     & ---                       & 67.6\% & 49.1\% & $-18.5$ \\
duration  & duration-noise scale 1.6  & 53.7\% & 20.4\% & $-33.3$ \\
rate      & speaking rate 1.35        & 61.1\% & 31.5\% & $-29.6$ \\
\bottomrule
\end{tabular}
\end{table*}

\noindent The gap here is control minus repeated: every VITS arm is
negative, repeated-side median relative error exactly zero in all
three, and both perturbations lower the control side further, a judge
effect silent on repetition; stock defaults were not load-bearing, and
the report is the sign, not the size. The null also does the judge's
work: the same repeated words, same test set and same recogniser
produce no gap, so whatever CTC blank-collapse costs, it does not
create the dissociation.

\noindent\textbf{What plausibly separates them.} VITS predicts a
duration per input token, so ``how many'' is carried structurally by
the input; F5-TTS estimates one total duration and denoises the whole
mel in parallel, so it must represent how much speech to produce, the
burden an autoregressive decoder carries in its state. A hypothesis,
not a result: formed after seeing the two baselines, resting on $n=2$
systems, untested.

\paragraph{The intervention: supplying the duration.} Handing F5-TTS
the correct total duration repairs the deficit below $k=12$ and does
nothing at all above: the paper's one intervention on the proposed
mechanism. The supplied value is the median of F5-TTS's own control
renderings at the same $k$ plus the measured reference-clip length,
never a function of the repeated item's output. Table~\ref{tab:f5oracle}
gives the split by $k$.

\begin{table*}[tp]\centering
\caption{F5-TTS handed the oracle total duration: exactly-right rate and median relative error, free against given, by $k$.}
\label{tab:f5oracle}
\begin{tabular}{rrrrr}
\toprule
$k$ & free exact (\%) & given exact (\%) & free median & given median \\
\midrule
6 & 46.7 & 80.0 & -0.167 & +0.000 \\
8 & 40.0 & 73.3 & -0.125 & +0.000 \\
12 & 33.3 & 40.0 & -0.083 & -0.083 \\
16 & 20.0 & 26.7 & -0.062 & -0.062 \\
24 & 6.7 & 0.0 & -0.167 & -0.125 \\
32 & 6.7 & 0.0 & -0.094 & -0.125 \\
\bottomrule
\end{tabular}
\end{table*}

Below $k=12$ exact rises from $43.3\%$ to $76.7\%$ and median error
goes to zero, the split the main text quotes; at or above, the gain is
$0.0$ points: given the correct total length, the model still cannot
place thirty-two repetitions inside it. The mild end is substantially
a duration-estimation failure; the severe end, where the AR panel's
deficit is worst, is not. One model, $n=15$ per cell, one seed triple,
a median rather than an item-specific target: a direction, not an
effect size.


\section{State instrumentation}
\label{sec:instrumentation}

The panel is the main text's six checkpoints; per-layer hidden states
were captured for every generation the state-space sections analyse, at
a per-checkpoint probe density of every layer to every fourth
($P=9$--$16$ probed layers). States were captured by
one teacher-forced forward pass over the realised token sequence
(Llasa, XTTS-v2) or read out inside the generation loop itself
(Qwen3-TTS), position $t$ carrying, by causality, the exact state that
produced token~$t$. Instrumented generation was validated equivalent to
normal generation before any measurement was interpreted: sampling is
byte-for-byte identical, and captured inputs reproduce incremental
generation. A seventh registry entry, XTTS-v2 with repetition penalty
disabled, is an ablation of that row (Section~\ref{sec:penalty}),
excluded by construction from every panel-pooled statistic.

\noindent\textbf{The count probe.} Ridge regression predicts $\log_2 k$
from mean-pooled hidden states under leave-one-template-out
cross-validation, penalty $\alpha=1.0$ at every layer, item family and
checkpoint, never swept or tuned. No layer is fixed in advance: per
checkpoint, item family and window (\emph{early}: mean state over the
trajectory's first third; \emph{late}: final third), the probe is fit
at all $P$ probed layers and the argmax of cross-validated $R^2$
reported. Selected-layer leave-one-template-out $R^2$ spans
$0.18$--$0.71$ across the $24$ checkpoint--family--window cells
($n{=}54$ items each); these fits are the ones behind
Section~\ref{sec:alternatives}'s retention ratios.

\section{The contraction premise, measured}
\label{sec:contraction}

This section expands the main text's negative result. The candidate
mechanism held that softmax attention over $k$ near-identical copies
flattens toward uniformity, so the decoder meets the same conditioning at
every repetition boundary and the boundary-to-boundary state map is
effectively autonomous; were that map a contraction (largest singular
value $q<1$), states would converge geometrically and no Lipschitz
readout, the stop head included, could separate counts past an explicit
horizon (the implication itself is machine-verified in the release). The hypothesis is quantitative: measure $q$. Here it is measured on
five checkpoints, three Llasa~\cite{ye2025llasa} and two
Qwen3-TTS~\cite{hu2026qwen3tts}, and rejected.

\noindent\textbf{Two estimators failed first.} The boundary-distance
estimator fit $\log d_m = a + m\log q$ to boundary-state distances
located by a text-attention centroid; its one assumption, monotone
centroid motion, is false (${\approx}51\%$ of steps advance, a random
walk), and it returned $\hat q\approx1.00$ $[0.99,1.02]$ everywhere it
ran. Its boundary-free replacement, windowed dispersion converted to a
per-repetition rate, put $\hat q$ below $1$ on the identical
trajectories: it measures typical spread, not the worst-direction
Lipschitz constant the horizon requires.

\noindent\textbf{The estimator behind the reported $q$.}
Teacher-forcing the generated tokens makes the window-to-window map a
deterministic, differentiable endomorphism; $q$ is its largest singular
value under a per-layer root-mean-square metric, required to survive a
sweep over four lags and sub-stacks. Teacher-forcing deletes the token
channel, biasing $q$ \emph{toward} the premise; independent cache
perturbation biases against. Four pre-committed validation gates passed
on all five checkpoints (synthetic positive control, relative error
$3.7\times10^{-4}$; exactly zero backward map; Jacobian-vector products
against finite differences, ${\le}1.5\times10^{-3}$; Qwen replay
cosine ${\ge}0.99920$).

\noindent\textbf{The five-checkpoint panel.}
Table~\ref{tab:qpanel} reports the headline setting (lag $\tau$, whole
stack): $346$ items (repeated plus length-matched control), $15{,}243$
attempted measurements, $14{,}796$ valid; the $447$ excluded are accounted for
below, not silently dropped. $q < 1$ occurs in $0$ of
$346$ items on either arm; the recorded fraction below $1$ is exactly
$0.0$, both arms, every checkpoint. Pooled, $q$ runs $21.0$--$347.7$,
the main text's quoted range; the most favourable cell anywhere
(smallest bootstrap lower bound over all lags, sub-stacks and arms) is
Qwen3-TTS-0.6B's $3.35$, a panel-wide floor of $3.3$. Contraction was
never a live description of these decoders.

\begin{table*}[tp]
\centering
\caption{Largest singular value $q$ of the per-repetition map at the
headline setting (lag $\tau$, whole stack), per arm, with the paired
repeated-versus-control test. Checkpoints as in
the main text's panel table.}
\label{tab:qpanel}
\small
\begin{tabular}{@{}lrrp{2.8cm}p{2.8cm}r@{}}
\toprule
checkpoint & layers & $n$ pairs & $q$ repeated [95\% CI] & $q$ control [95\% CI] & paired $p$ \\
\midrule
Llasa-1B & 16 & 29 & 38.1 [32.9, 51.4] & 31.8 [27.3, 39.8] & 0.017 \\
Llasa-3B & 28 & 26 & 118.8 [101.9, 156.5] & 86.4 [77.2, 137.3] & 0.269 \\
Llasa-8B & 32 & 29 & 347.7 [258.1, 393.5] & 185.4 [142.1, 236.5] & 0.256 \\
Qwen3-TTS-0.6B & 28 & 44 & 21.0 [19.7, 24.2] & 21.9 [20.6, 23.2] & 0.931 \\
Qwen3-TTS-1.7B & 28 & 45 & 24.7 [23.1, 25.7] & 23.6 [21.4, 25.1] & 0.177 \\
\bottomrule
\end{tabular}
\end{table*}

\noindent\textbf{Both arms expansive, in all sixteen cells.} In all
sixteen (lag $\times$ sub-stack) cells, on every checkpoint, on
\emph{both} arms, the bootstrap CI for $q$ lies entirely at or above
$1$ and the fraction of items with $q < 1$ is $0.0$
(Table~\ref{tab:qcells}). The one significant paired contrast points the \emph{wrong} way
(Llasa-1B: repetition \emph{more} expansive than its control); the rest
are null (Table~\ref{tab:qpanel}). The premise fails on both arms:
contraction was never there for periodicity to break.

\begin{table*}[tp]
\centering
\caption{Range of $q$ over all sixteen (lag $\times$ sub-stack) cells,
per arm. Checkpoints as in the main text's panel table.}
\label{tab:qcells}
\small
\begin{tabular}{@{}lp{3.6cm}p{3.6cm}@{}}
\toprule
checkpoint & repeated: range over 16 cells & control: range over 16 cells \\
\midrule
Llasa-1B & [3.90, 78.10] & [3.50, 46.61] \\
Llasa-3B & [9.33, 182.80] & [7.60, 148.28] \\
Llasa-8B & [11.43, 499.13] & [12.28, 329.33] \\
Qwen3-TTS-0.6B & [3.35, 29.47] & [3.41, 28.23] \\
Qwen3-TTS-1.7B & [3.45, 37.83] & [3.35, 31.18] \\
\bottomrule
\end{tabular}
\end{table*}

\noindent\textbf{Two facts sharpen the refutation past replication.}
First, $q$ grows with scale inside Llasa: one family, same estimator,
same headline cell, $38.1 \to 118.8 \to 347.7$ from 1B through 8B, so a
``contraction mechanism specific to the larger checkpoints'' is
excluded in the rescuing direction. Second, Qwen3-TTS-1.7B, the one
checkpoint with zero median count error, is exactly as expansive as the
rest ($q = 24.7$ $[23.1, 25.7]$): whatever separates counting from
miscounting here, it is not $q$.

\noindent\textbf{Skip accounting.} Llasa-8B skipped $21.8\%$ of its
$1677$ attempted measurements ($236$ exceeding its $700$-token read
window, $129$ out of memory, reported distinctly); Llasa-3B skipped
$3.1\%$ ($70$ of $2286$, all window), Llasa-1B $0.4\%$, and both Qwen
checkpoints none. Every skipped anchor is recorded with its reason;
truncation is exact on a causal model (a shorter read window changes
which anchors form, not the values returned).


\section{Dilution at the item level, and probing past the ladder}
\label{sec:dilution}

Within each (model, $k$) cell, the flattest-attention items are counted
\emph{best}: the item-level reading of Lemma~1 fails with the sign
reversed. Flattening that destroys the ability to tell occurrences
apart predicts a \emph{negative} within-cell correlation between
attention flatness and count error (undercounting makes count error
negative). Measured over three flatness measures fixed in advance,
signed so higher is flatter and pooled by Fisher $z$ weighted by cell
size, all three run positive on repeated items with intervals excluding
zero, and controls sit at zero on all three (nothing to dilute;
Table~\ref{tab:flatness}).
\begin{table*}[tp]\centering
\caption{Within-cell correlation between attention flatness and count error, pooled by Fisher $z$; positive means flatter items are counted better.}
\label{tab:flatness}
\begin{tabular}{lrr}
\toprule
flatness measure & repeated & control \\
\midrule
largest single share   & +0.59 [+0.35, +0.75] & +0.01 [-0.34, +0.37] \\
block entropy          & +0.43 [+0.16, +0.65] & +0.17 [-0.20, +0.49] \\
deviation from uniform & +0.55 [+0.30, +0.72] & +0.05 [-0.31, +0.39] \\
\bottomrule
\end{tabular}
\end{table*}
Only the item-level reading falls: Lemma~1 still holds in aggregate
(near-uniform attention over repeated spans, entropy growing as
$\log k$), but dilution does not pick which generations fail. Reverse
causation stays open, a stalled decoder yielding exactly this peaked
profile; only intervening on attention could separate consequence from
cause.

\noindent\textbf{Probing past the horizon.} The contraction hypothesis
forbade a Lipschitz readout from separating counts past $N^\ast$,
saying nothing below (Section~\ref{sec:contraction}); every probe so
far was fitted on the main ladder, $k\le32$, where it permits success.
The $k=48\ldots128$ items were regenerated with hidden-state capture on
five checkpoints, three Llasa~\cite{ye2025llasa} and two
Qwen3-TTS~\cite{hu2026qwen3tts} (XTTS-v2's token ceiling censors every
$k\ge48$ item), with the same ridge probe and folds; twelve aperiodic
control items matched on $k$ and template give each checkpoint its own
control arm. Because $k=48\ldots128$ spans a third of the main ladder's
$\log_2$ range, low $R^2$ alone proves nothing; the constant predictor
is the yardstick (Table~\ref{tab:probeladders}).
\begin{table*}[tp]\centering
\caption{Ridge-probe decoding of the requested count against the constant predictor, below the horizon ($k\le32$) and past it ($k=48\ldots128$).}
\label{tab:probeladders}
\begin{tabular}{llrrr}
\toprule
checkpoint & arm & probe MAE & constant MAE & $R^2$ \\
\midrule
Llasa-1B       & repeated $k=2\ldots32$    & 0.63 & 1.12 & $+0.62$ \\
Llasa-3B       & repeated $k=2\ldots32$    & 0.86 & 1.12 & $+0.41$ \\
Llasa-8B       & repeated $k=2\ldots32$    & 0.75 & 1.12 & $+0.52$ \\
Qwen3-TTS-0.6B & repeated $k=2\ldots32$    & 0.79 & 1.12 & $+0.47$ \\
Qwen3-TTS-1.7B & repeated $k=2\ldots32$    & 0.79 & 1.12 & $+0.47$ \\
\midrule
Llasa-1B       & repeated $k=48\ldots128$  & 0.50 & 0.50 & $-0.02$ \\
Llasa-3B       & repeated $k=48\ldots128$  & 0.49 & 0.50 & $-0.02$ \\
Llasa-8B       & repeated $k=48\ldots128$  & 0.43 & 0.50 & $+0.19$ \\
Qwen3-TTS-0.6B & repeated $k=48\ldots128$  & 0.37 & 0.50 & $+0.34$ \\
Qwen3-TTS-1.7B & repeated $k=48\ldots128$  & 0.45 & 0.50 & $+0.15$ \\
\midrule
Llasa-1B       & control $k=48\ldots128$   & 0.43 & 0.50 & $+0.26$ \\
Llasa-3B       & control $k=48\ldots128$   & 0.41 & 0.50 & $+0.33$ \\
Llasa-8B       & control $k=48\ldots128$   & 0.29 & 0.50 & $+0.56$ \\
Qwen3-TTS-0.6B & control $k=48\ldots128$   & 0.47 & 0.50 & $+0.10$ \\
Qwen3-TTS-1.7B & control $k=48\ldots128$   & 0.44 & 0.50 & $+0.20$ \\
\bottomrule
\end{tabular}
\end{table*}
All five decode the count below the horizon. Past it, the probe-MAE
ratio against the constant predictor decides, ratios within $\pm0.02$
of $1$ called indistinguishable rather than rounded either way
(Table~\ref{tab:probeverdicts}).
\begin{table*}[tp]\centering
\caption{Past-horizon verdicts: probe-to-constant MAE ratios with the $\pm0.02$ tie band, and the length-only control on the repeated arm.}
\label{tab:probeverdicts}
\begin{tabular}{lrlrrr}
\toprule
 & \multicolumn{3}{c}{MAE ratio vs.\ constant} & \multicolumn{2}{c}{MAE, repeated arm} \\
\cmidrule(lr){2-4}\cmidrule(lr){5-6}
checkpoint & rep. & reading & ctl. & length-only & state probe \\
\midrule
Llasa-1B       & 1.01 & indistinguishable & 0.86 & 0.61 & 0.51 \\
Llasa-3B       & 0.99 & indistinguishable & 0.83 & 0.56 & 0.49 \\
Llasa-8B       & 0.86 & recovers the count & 0.59 & 0.45 & 0.43 \\
Qwen3-TTS-0.6B & 0.74 & recovers the count & 0.95 & 0.30 & 0.37 \\
Qwen3-TTS-1.7B & 0.90 & recovers the count & 0.89 & 0.36 & 0.45 \\
\bottomrule
\end{tabular}
\end{table*}
Every control arm sits below $1$, none in the tie band: where the count
is lost, narrowness of the target range is ruled out on that
checkpoint's own data. Under the strict threshold the contraction
account's prediction holds in 1 of 5; by effect size, 2 of 5 show
nothing recoverable past the horizon and 3 clearly do. We quote 1 of 5,
the less favourable reading. And the last two columns void the Qwen
recoveries: a predictor given only log trajectory length, no hidden
states, beats the state probe on both Qwen checkpoints and loses on all
three Llasa (repeated arm, $k=48\ldots128$, same folds).
Budget-truncated trajectories are far commoner on the repeated arm
(1--6 of 12 items per checkpoint, against 0--1 on the matched control;
worst, 6 of Qwen3-TTS-0.6B's 12), and truncated length is not
independent of $k$: babble length alone carries $k$, no count read from
state. The two Qwen recoveries are voided; the three Llasa probes beat
the baseline, and their representational reading stands.

\section{Causal interventions on the count}
\label{sec:causal}

No causal conclusion stands: the intervention instrument fails its own
positive control. Full-residual activation patching, overwriting the
residual stream at 128 consecutive positions with a donor's states, is
too disruptive to interpret: states from an \emph{unrelated} item move
the rendered count $1.34\times$ as much as a donor that actually
differs in $k$, and even a length-matched same-$k$ control donor drives
$44.4\%$ of its runs into the token budget. The rank-1 replacement,
transferring only the probe's count coordinate, removes the damage and
returns equivalence bounds (Table~\ref{tab:causal-bounds}); but its
positive control, the identical write applied to a carrier direction
Qwen3-TTS-1.7B decodes perfectly (leave-one-template-out sign accuracy
$1.00$), moves the output by at most $1.3\%$ of a full carrier transfer
at the published magnitude, and on every readout tested the informative
write is statistically indistinguishable from a pure sampler re-roll
($p \ge 0.38$). A null from an intervention transferring nothing along
\emph{any} direction says nothing about the count direction.

\begin{table}[htbp]
\centering
\caption{Rank-1 count-coordinate patching on four checkpoints in two
families~\cite{ye2025llasa,hu2026qwen3tts}, $2{,}178$ scored runs,
thresholds pre-committed before any audio. P1 ratio:
unrelated-to-informative disruption. Worst cell: largest median shift
across donor kinds as a fraction of a full donor-to-receiver transfer
(noise-floor bar: below $25\%$). Bound: $90\%$ TOST equivalence bound
on the informative cross-$k$ cell in the same units, the smallest
excludable effect.}
\label{tab:causal-bounds}
\small
\begin{tabular}{@{}lrrrl@{}}
\toprule
checkpoint & P1 ratio & worst cell & bound & verdict \\
\midrule
Qwen3-TTS-0.6B & 1.38 & $14\%$ & $9\%$  & readable null \\
Qwen3-TTS-1.7B & 1.12 & $9\%$  & $6\%$  & readable null \\
Llasa-1B       & 0.57 & $25\%$ & $20\%$ & readable null \\
Llasa-8B       & 0.90 & $42\%$ & ---    & cannot tell \\
\bottomrule
\end{tabular}
\end{table}

Llasa-8B fails decisively: worst cell $42\%$ of full transfer, a
same-$k$ donor carrying exactly the receiver's count disrupting as much
as a cross-$k$ one ($0.246$ against $0.217$ median absolute shift),
unchanged at doubled seeds; an arm where the right answer disrupts like
the wrong one measures foreign state, not different count. The verdict,
\emph{cannot tell}, is not \emph{no}, and the main text says so. What
survives is methodological: rank-1 probe-direction patching at this
magnitude is no working instrument on these decoders, and a causal null
from this family is only as good as its positive and disruption
controls; ours initially lacked both.

\end{document}